%% file: SAMBA-Arxiv.tex
\documentclass{article}
\usepackage[paper=letterpaper,centering,margin=1.0in]{geometry}

\usepackage{amssymb,amsmath,amsthm}
\usepackage{authblk}
\usepackage{graphicx}

\newtheorem{theorem}{Theorem}
\newtheorem{corollary}{Corollary}
\newtheorem{lemma}{Lemma}
\newtheorem{remark}{Remark}
\newtheorem{proposition}{Proposition}
\newtheorem{definition}{Definition}
\DeclareMathOperator*{\argmax}{argmax}

\usepackage{bm}
\usepackage{xcolor}
\usepackage[ruled,vlined]{algorithm2e}
\usepackage{hyperref}

\newcommand{\Beginproof}{\begin{proof}}
\newcommand{\Endproof}{\end{proof}}

\usepackage[square,numbers]{natbib}
\begin{document}

\title{Regret Analysis of a Markov Policy Gradient Algorithm for Multi-arm Bandits}

\author{Denis Denisov}
	\author{Neil Walton}
\affil{University of Manchester\\
\small{\texttt{\{denis.denisov, neil.walton\}@manchester.ac.uk}}}

\maketitle

\abstract{
We consider a policy gradient algorithm applied to a finite-arm bandit problem with Bernoulli rewards.
We allow learning rates to depend on the current state of the algorithm, rather than use a deterministic time-decreasing learning rate. 
The state of the algorithm forms a Markov chain on the probability simplex. 
We apply Foster-Lyapunov techniques to analyse the stability of this Markov chain.
We prove that if learning rates are well chosen then
the policy gradient algorithm is a transient Markov chain and the state of the chain converges on the optimal arm with logarithmic or poly-logarithmic regret. 
}

\maketitle

\input{INTRODUCTION}

\input{MODEL}

\input{THEOREM_1}

\input{THEOREM_av}

\input{SIMULATIONS}
\input{CONCLUSION}

\section*{Acknowledgments.}
Bandits is a new area for both authors. So we are grateful to Tor Lattimore for references, comments and suggestions on the positioning of this work. 
We are grateful to anonymous referee who suggested the average version of SAMBA considered in Theorem 3. 

\bibliography{SAMBA} 

\appendix
\input{APPENDIX}

\input{THEOREM_3}

\input{THEOREM_3e}

\input{REGULAR_VARIATION}

\end{document}

%% file: INTRODUCTION.tex
\section{Introduction.}

In a multi-armed bandit problem an algorithm must sequentially choose among a set of actions, or \emph{arms}. When selected, an arm produces a reward that is random with an unknown mean. 
The objective is to maximize cumulative reward over time. A good algorithm must efficiently explore the set of arms determining enough information so that it can concentrate selection on the arm with the highest reward. 
The performance of an algorithm is typically measured in terms of its regret, which is the difference between the cumulative reward of the optimal arm and the cumulative reward of the algorithm. As we will review shortly, there are a variety of algorithms that can be applied to multi-arm bandit problem and that have low regret. 
One class of algorithms, however, that are not well-understood are policy gradient algorithms. 

Policy gradient algorithms are extensively applied in reinforcement learning.
Multi-arm bandit problems can be viewed as a special case of reinforcement learning, and often results initially proved in the bandit setting are then later developed for more general reinforcement learning problems.
Policy gradient algorithms parametrize probabilities and maximize rewards by applying stochastic gradient ascent to the probability of selecting a given arm (or action). 
This contrasts value function methods which aims to directly estimate the reward of each arm, either by randomly exploring arms or by forming confidence bounds on the estimated reward. 
The theory of value function methods is much more developed than policy gradient methods, both in reinforcement learning and in multi-arm bandit problems.
A good example of a policy gradient algorithm in the bandit setting is given in the book of Barto and Sutton \cite{sutton2018reinforcement}. There William's REINFORCE algorithm \cite{williams1992simple} is specialized to the bandit setting and with a fixed learning rate and simulation results find it to have good performance.

Despite good empirical performance, regret bounds for policy gradient algorithms are scarce, even for bandit problems. Only recently has substantial progress been made, and this is for the deterministic analogues of these randomize policies \cite{agarwal2019optimality,bhandari2019global,mei2020global}. The only work we are aware in the stochastic case is the recent paper of \cite{zhang2020sample}. This establishes an order $O(T^{5/6} (\log T)^{5/2})$ regret bound for the REINFORCE algorithm. 
Since statistical consistency results and stochastic regret bounds do not, in general, exist in prior work, one task of this article is to prove almost sure convergence and a regret bound for a policy gradient algorithm. This is amongst the first sub-linear regret bounds for a policy gradient algorithm, albeit, for a simple somewhat canonical bandit setting: a finite-arm bandit problem with Bernoulli distributed rewards. 

Another important aspect of this article is to investigate the use of Markov chain tools to analyze stochastic approximation and optimization. Over the last decade, researchers have developed a much clearer understanding of the finite time error of stochastic approximation, online optimization and bandit problems. 
Such analysis typically requires a deterministic decreasing step-size.
More recently, there has been an increased interest in stochastic approximation where the step-size is fixed \cite{babichev2018constant,dieuleveut2017bridging}. In this case, the stochastic recursion is a Markov process and convergence is understood by analysing ergodic behaviour of this process. In this paper, we also choose step-sizes so that our algorithm evolves as a Markov chain. In contrast to prior work, we analyse transient rather than ergodic properties. A careful analysis of the rate of transience gives our regret bounds. 

A Markov chain policy gradient algorithm has a design that is conceptually different to mainstream bandit algorithms: the algorithm estimates probabilities in a time-independent manner, rather than estimate rewards in a time-dependent manner.
Consequently, our regret analysis requires different mathematical tools.
Our analysis relies on Foster-Lyapunov results for the convergence of our algorithm \cite{meyn2012markov} as well as Markov chain coupling techniques.
This Markov chain approach is new both in the context of Bandit problems and in the context of Policy Gradient algorithms. We consider a variant of our algorithm that operates on average rewards. We demonstrate mathematically and empirically that this algorithm also has good performance. 
One aim is that the mathematical results and methods applied in this paper can be both refined and generalized to understand the performance of different policy gradient algorithms in a wide variety of settings. 

The results that we prove apply to a specific policy gradient algorithm which we call SAMBA: Stochastic Approximation Markov Bandit Algorithm. We first prove that logarithmic regret bounds are achievable for a suitably small learning rate which depends on the gap between mean reward of each arm. We then modify this algorithm to remove the dependence on the gap and prove a $O(\log(T)^2)$ regret bound and a $O(\log( T) \cdot \log (\log(T)))$ regret bound.
At a technical level, it remains to be proven that $O(\log(T))$ bounds are achievable for policy gradient methods as they are for value function methods. This is a significant open problem. Nonetheless, it is also important step that both consistency and poly-logarithmic regret bounds are achievable for policy gradient algorithms. We focus on bandit problems in this paper but 
an important future research direction is to extend these methods to reinforcement learning.
%
%

\subsection{Further Literature.} 
There are a number of excellent texts that overview multi-arm bandit problems from different perspectives in a variety of settings \cite{bubeck2012regret,gittins2011multi,lattimore2018bandit,hazan2016introduction}. A recent review of application areas is \citep{bouneffouf2019survey}.
A list of the most popular algorithms for stochastic multi-arm problems with a finite number of arms is: 
Upper Confidence Bound (UCB) \citep{agrawal_1995,auer2002finite}, Exponential Explore Exploit (Exp3) \citep{auer1995gambling,seldin2012evaluation}, Thompson Sampling \citep{thompson1933likelihood,Kaufmann2012,Agrawal2012}, Mirror Descent / Regularized-Follow-the-Leader by \cite{abernethy2009competing}, Explore and Commit \citep{anscombe1963sequential}, $\epsilon$-Greedy \citep{sutton2018reinforcement}.
Each of these methods maintains an estimate for the expected reward of each arm. Typically algorithms maintain time dependent parameters that are used in order to concentrate selection on the best arm.


A different approach is to apply a policy gradient algorithm. 
As discussed, a policy gradient algorithm directly applies stochastic gradient optimization to the probability of selecting each arm.
Rewards are not explicit estimated, instead the probability of selecting the optimal arm is the object of interest.
Methods of this type were first introduced by Williams \cite{williams1992simple} for reinforcement learning problems. 
Bandit algorithms can be viewed as an important special case of reinforcment learning. 
A good example of this approach to bandit problems is given in the text of Barto and Sutton \cite{sutton2018reinforcement}.
Here William's original REINFORCE algorithm is applied to the multi-arm bandit problem. This Gradient Bandit Algorithm (GBA) applies a softmax function, and under this parametrization a gradient ascent algorithm with importance sampling is applied. 
Regret bounds for REINFORCE both in bandit problems and general reinforcement learning have not been established. Progress on deterministic analogous of policy gradient algorithms is underway \cite{agarwal2019optimality,bhandari2019global,mei2020global}. However, results for the stochastic systems and regret bounds are still not well understood. In this paper, we make progress on this problem albeit in the more specialized setting of bandit problems.  

An interesting feature of Barto and Sutton's Gradient Bandit Algorithm is that good performance can be found with a fixed learning rate. 
If the learning rate is fixed or only dependent on the current state of algorithm, then the algorithm evolves as a Markov chain.
We apply a learning rate that also yields a time-homogeneous Markov process.
There are certain conceptual advantages to this approach, for instance, the learning processes does not need to be reset and the algorithm does not require a notation of how much time has elapsed in the learning process.

 Recent works consider Markov analysis of stochastic gradient descent with fixed learning rate, see \cite{babichev2018constant,dieuleveut2017bridging}. Similar approaches have also been applied to reinforcement learning, see \cite{beck2012error,srikant2019finite}. In these prior works the Markov chain is recurrent and the stationary distribution of the error about the optimum is analysed. The error does not vanish over time. In contrast, the
Markov process we consider is a {transient} Markov chain, and the error decays at rate $O(1/t)$ to the correct solution. 

The state of the SAMBA algorithm is a Markov chain on the probability simplex. Our proof applies Foster-Lyapunov techniques for continuous state-space Markov chains \cite{meyn2012markov}. 
In the operations research literature, there is a well developed theory of recurrence and transience of random walks in polytopes which helps to inform our analysis and our choice of Lyapunov functions \cite{dupuis1994lyapunov,kushner2013heavy}. The Markov processes considered here are necessarily close to the threshold between recurrence and transience. Here essential criteria and techniques were initiated by \cite{LAMPERTI1,LAMPERTI2}. 
See~\cite{denisov2016edge} and~\cite{denisov2020renewal} for a recent review of methods.


\subsection{Organization.} The remainder of the paper is organized as follows. In Section \ref{sec:Prelim}, we present the SAMBA algorithm and our main results, namely, Theorem \ref{main_theorem} and Theorem \ref{3rd_theorem}. 
In Section \ref{sec:model}, we more formally describe the multi-arm bandit model, the policies considered and we define additional mathematical notation required for the proofs.  
In Section \ref{sec:Proofs}, we prove Theorem \ref{main_theorem}. The proof of Theorem  \ref{3rd_theorem} follows a very similar argument. For this reason the proof of Theorem \ref{3rd_theorem} is presented in Appendix \ref{sec:3rd_proof}. A further extension, Theorem \ref{4th_theorem}, is also given in the appendix. A simulation study is provided in Section \ref{Simulations}. This confirms a number of characteristics discussed within the proofs, and, also, provides comparison with some multi-arm bandit algorithms. We then conclude the paper in Section \ref{sec:Conc}.

\section{Preliminaries.} \label{sec:Prelim}

In this section, we give a heuristic derivation the policy gradient algorithm, SAMBA, and explain why logarithmic regret bounds might be expected for this algorithm. Then, after defining the algorithm, we present the main results of the paper. Afterwards, we perform a literature review of relevant works. 

\subsection{Heuristic Motivation.}\label{Heuristic}
We describe the algorithm analyzed in this paper. What follows is a heuristic derivation. 
A more formal description and proofs are given subsequently.

We consider a multi-arm bandit problem with arms $a\in \mathcal A$. The reward from arm $a$ is given by a random variable $R_a$ with values in $\{0,1\}$ and with mean $r_a$. We denote the optimal arm by $a^\star$, that is $r_{a^\star} > r_a$ for all $a\neq a^\star$.
We let $N =|\mathcal A| $ be the number of arms. 
We let $\Delta_a := r_{a^\star}-r_a$ and we let $\Delta:= r_{a^\star}-\max_{a \neq a^\star} r_a$. We assume $\Delta >0$. 
We let $p_a$ be the probability of playing arm $a$ and $I_a$ be the indicator function that arm $a$ is played. The deterministic analogue of minimizing regret is the following linear program:
\begin{equation*}
\text{minimize}\quad 
\sum_{a\in \mathcal A} p_a (r_{a^\star} - r_{a}) \quad \text{subject to} \quad \sum_{a\in \mathcal A} p_a = 1 \quad \text{over} \quad p_a \geq 0,\quad a\in \mathcal A.
\end{equation*}
We now discuss how we form a stochastic gradient descent rule on this optimization.
Gradient descent would perform the update $p_a \leftarrow p_a + \gamma (r_a-r_{a^\star})$, for $a\neq a^\star$. However, since the mean rewards $r_a, a\in \mathcal A,$ are not known, a stochastic gradient descent must be considered: $p_a \leftarrow p_a + \gamma (R_a-R_{a^\star})$, $a\neq a^\star$. 
Also, the optimal arm is unknown. So instead of $a^\star$, we let $a_\star$ be the arm for which $p_a$ is maximized and, in place, consider the update $p_a \leftarrow p_a + \gamma (R_a-R_{a_\star})$, $a\neq a_\star$.
Since the reward from only one arm can be observed at each step, we apply importance sampling: 
\begin{equation*}
p_a \leftarrow p_a + \gamma 
\Big( \frac{R_aI_a}{p_a}-\frac{R_{a_\star}I_{a_\star}}{p_{a_\star}} \Big),\quad 
\text{for }a\neq a_\star\, .
\end{equation*}
Alternatively, we can record the average reward obtained by each arm to give the update
\begin{align*}
  p_a \leftarrow p_a + \gamma (\hat r_a - \hat r_{\star}),\qquad \text{for }a\neq a_\star\, .
\end{align*}
 where $\hat r_a$ is the empirical mean reward of arm $a$ and $\hat r_{\star}= \max_a \hat r_a$.
Both of the above, gives a simple recursions for a multi-arm bandit problem.

Finally, let's consider the learning rate $\gamma$. Again, consider the gradient descent update $p_a \leftarrow p_a + \gamma (r_a-r_{a^\star})$, for $a\neq a^\star$.
Notice if we let 
$
\gamma = \alpha p_a^2	
$
then the gradient descent algorithm approximately obeys the following differential equation:
$$
\dot p_a = -\alpha p_a^2 (  r_a - r_{a^\star})\, ,
$$
whose solution is $$p_a(t) = \frac{p_a(0)}{1+\alpha \Delta_ap_a(0) t}\, .$$
The regret, $\mathcal R\!g(T)$, which is the accumulated difference between the optimal reward and the algorithm can be analysed as follows:
\begin{align*}
\mathcal R\! g(T) = \int_0^T \sum_a (r_{a^\star} -r_a) p_a(t) dt 
&
\leq  
\int_0^T \sum_{a\neq a^\star} p_a(t) dt 
\\
&
\leq 
\sum_{a\neq a^\star} p_a(0)
\int_0^T \frac{1}{1+\alpha  p_a(0)\Delta_a t} dt
\\
&
= 
\sum_{a \neq a^\star}
\frac{1}{\alpha\Delta_a}
\log ( 1+ \alpha p_a(0) \Delta_a T  )
\sim 
\sum_{a\neq a^\star} \frac{1}{\alpha \Delta_a} \log T
.
\end{align*}
The regret of the algorithm grows as the sum of these probabilities, see Lemma \ref{Regret_Lemma}.
This suggest a learning rate of $\gamma = \alpha p_a^2$, applied to each $a$, gives a logarithmic regret. Theorem \ref{main_theorem} is the formal version of this argument. 

\smallskip
\subsection{Algorithm and Main Results.} 
To summarize, our first algorithm takes data 
$(R_aI_a : a\in \mathcal A)$ and performs stochastic approximation update:
\begin{equation}
\label{p_gamm_update}
p_a \leftarrow p_a + \alpha p_a^2 
\Big( \frac{R_aI_a}{p_a}-\frac{R_{a_\star}I_{a_\star}}{p_{a_\star}} \Big),\quad 
\text{for }a\neq a_\star\, .
\end{equation}
We call the algorithm SAMBA: Stochastic Approximation Markov Bandit Algorithm.
Over time, the probabilities 
$(p_a(t) :a\in\mathcal A)$ 
are a Markov chain. The algorithm directly applies a stochastic gradient descent on probabilities of actions. Thus, it is a policy gradient algorithm.
 Pseudo-code is given above. 
For $\alpha\in (0,1)$, these probabilities remain in the probability simplex. (This is proven in Lemma \ref{lemma3}).

\begin{figure}
	\centering
	\includegraphics[width=1.\textwidth]{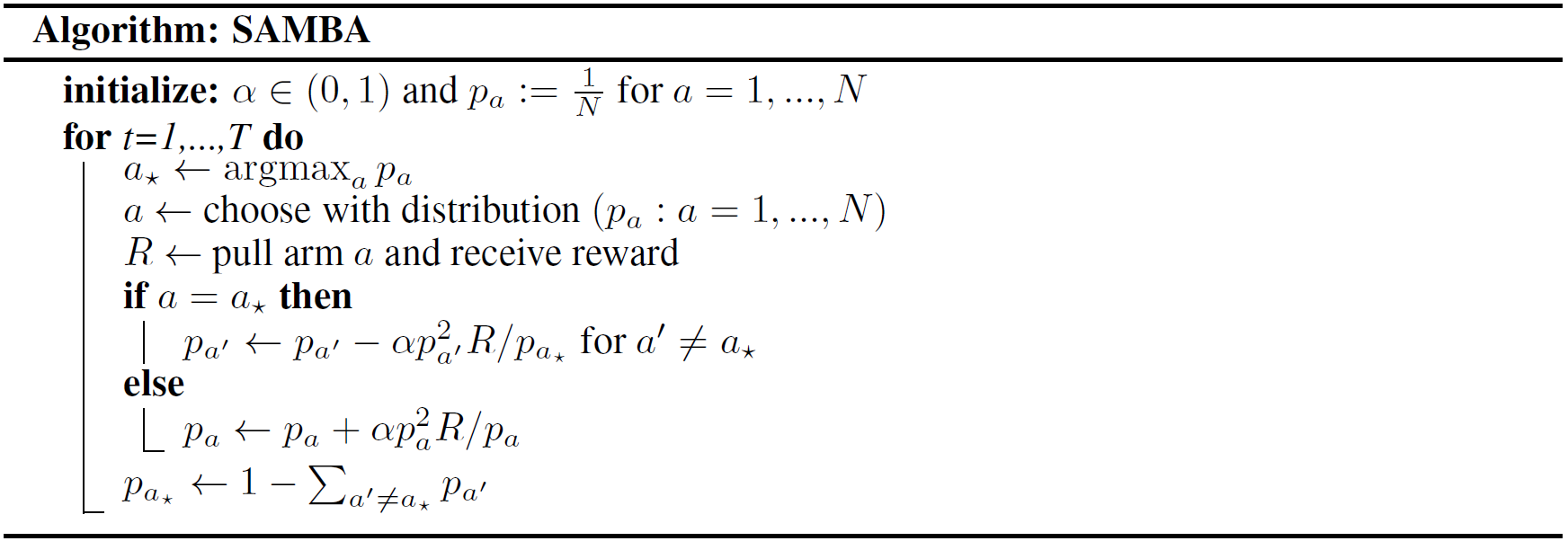}
\end{figure}



We prove that SAMBA has logarithmic regret for sufficiently small learning rates. We use $\mathcal R\!g(T)$ to denote the regret at time $T$. This is the difference between the cumulative reward of the optimal arm and the cumulative reward of the algorithm, and is defined in \eqref{Regret} in Section \ref{sec:model}. 
The following is our main result for SAMBA. 

\begin{theorem}\label{main_theorem}
	If $\alpha$ is such that 
	\begin{equation}\label{alpha_cond}
	\alpha < \frac{\Delta}{r^\star - \Delta}
	\end{equation}
	then the SAMBA process $(p_a(t) : a\in\mathcal A)$, $t\in \mathbb Z_+$, is a Markov chain such that, with probability $1$,
	$
	p_{a^\star} (t) \xrightarrow[]{} 1
	$
	, $t\rightarrow\infty$
	and 
	\begin{equation}\label{regret_bound}
	\mathcal R\! g (T) 
	\leq 
	\frac{N}{\alpha\Delta }\log T + Q\, ,
	\end{equation}
	where 
	$
	Q:= \sum_{t=0}^\infty 
	\mathbb P
	\big(
	p_{a^\star} (t) \leq \frac{1}{2}
	\big) < \infty \, .
	$
\end{theorem}
We will discuss conditions shortly, but we may prefer a result where the condition on $\alpha$ does not depend on the gap, $\Delta$. For this reason, we let $\alpha$, applied to arm $a$, be a function of the probability of selecting arm $a$. In this way we can decrease $\alpha$ as $p_a$ goes to zero.
We prove the following theorem:

\begin{theorem}
	\label{3rd_theorem}
	If $\alpha: [0,1] \rightarrow [0,1]$ is such that
	\begin{equation}\label{alpha_2nd_cond_III}
	\alpha(p_a) = \frac{\beta}{\log ( e - \log p_a)}
	\end{equation}
	for $\beta\in(0,1]$ then the SAMBA process $(p_a(t) : a\in\mathcal A)$, $t\in \mathbb Z_+$, is a Markov chain such that, with probability $1$, 
	$
	p_{a^\star} (t) \rightarrow {} 1
	$ as $t\rightarrow\infty$
	and
	\begin{equation*}
	\mathcal R\! g(T)
	\leq 
	\frac{N}{\beta\Delta}
	\log T
	\cdot \log (e+\log ( T))
	+
	Q
	+
	1 \, .
	\end{equation*}
\end{theorem}
The proof of Theorem \ref{3rd_theorem} requires modification of some results used to prove Theorem \ref{main_theorem}, but the main structure of the proof is the same. We detail the proof in Section \ref{sec:3rd_proof}. 
Furthermore, it is possible to achieve an upper bound of the regret function, 
which is  $\log(T)$ multiplied by an arbitrarily slowly increasing (to infinity) function. This is Theorem \ref{4th_theorem}., which is stated and proven in Appendix~\ref{sec:4th_theorem}.

The second algorithm performs the update
\begin{align*}
  p_a \leftarrow p_a + \alpha p_a^2 ( \hat r_a - \hat r_{a_\star} )\, , \qquad a \neq a_\star
\end{align*}
where $a_\star \in \argmax \hat r_a$.
We call this SAMBA with averaging. Pseudo code is given above.

\begin{figure}
	\centering
	\includegraphics[width=1.\textwidth]{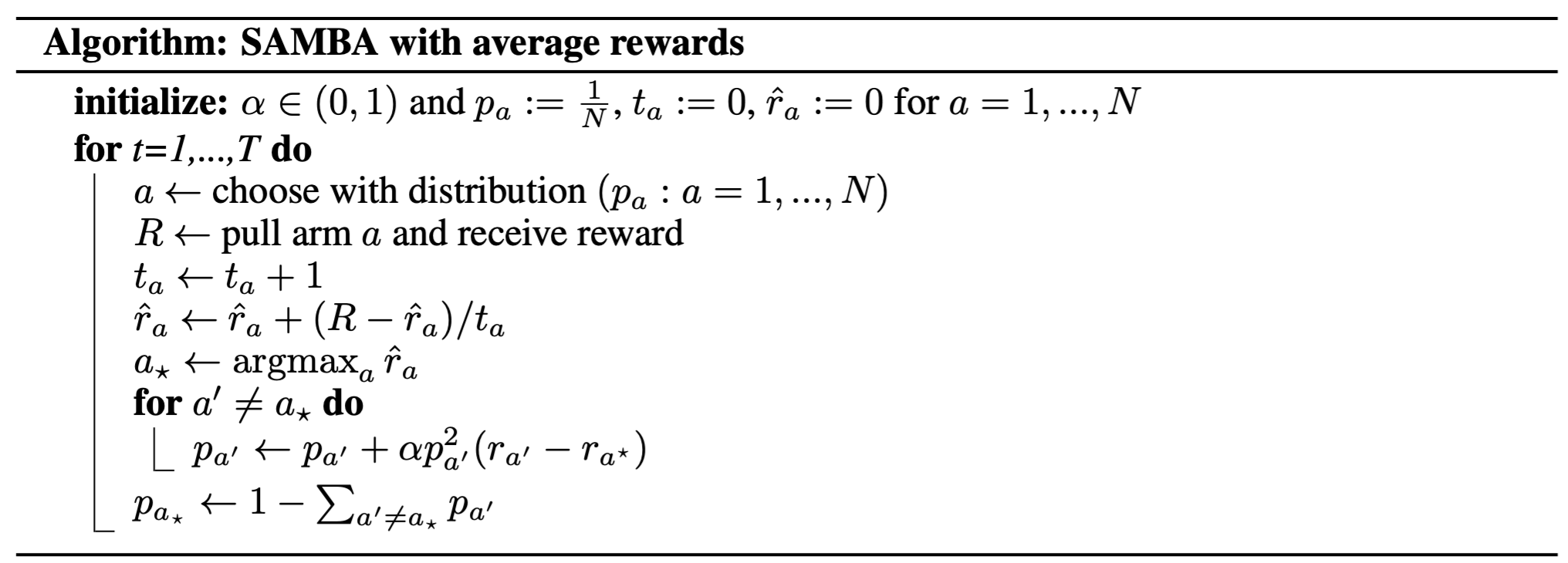}
\end{figure}

Similar to Theorem \ref{main_theorem}, logarithmic growth in regret is achievable provided the learning rate is sufficiently small. Specifically, the following result can be proved. 

\begin{theorem}\label{ThrmAv}
There exists a value $\alpha^\star \geq \frac{\Delta^2}{72}$ and a constant $Q$ such that for all $\alpha < \alpha^\star$ 
\begin{align*}
\mathcal R \! g(T)  \leq 
Q
+
 \frac{3 N }{\alpha \Delta} \log \left( 1+ \alpha \Delta T / 3 \right) \, .
\end{align*}
\end{theorem}

Like with Theorem \ref{3rd_theorem}, it is likely that choosing a state dependent learning rate will allow for a gap independent bound, perhaps at the cost of an increasing rate in the regret bound. However, since we now allow ourselves to estimate rewards, we can directly estimate $\Delta$ and use this to chose an appropriate $\alpha$. Further we note that the resulting probabilities are no longer a time homogeneous Markov chain. Thus a somewhat more standard bandit analysis can be applied. For this reason the proof in the Markovian case is perhaps more interesting. Nonetheless, the resulting averaging algorithm has very good performance on the simulations that we conducted. More broadly, the result is an interesting addition to Theorem \ref{main_theorem} as it suggests that the basic policy gradient algorithm is robust to different methods of estimating rewards (e.g. importance sampled rewards, which have a high variance, and averaged rewards with a much lower variance).

\medskip
\noindent 
\textbf{Discussion on Conditions.}
In Theorem \ref{main_theorem}, if $\alpha$ is chosen sufficiently small then regret is logarithmic.
The threshold that is necessary and sufficient  for logarithmic regret is not known. The condition \eqref{alpha_cond} implies that 
\[
\alpha < \frac{\Delta}{r^\star}.
\]
The proof  suggests that even if $\alpha$ is chosen too big (so that
condition \eqref{alpha_cond} does not hold) then the algorithm will still converge to arms within this factor of the optimal arm, e.g. if you want an accuracy of $99\%$ choose $\alpha=0.01$.  The condition on $\alpha$ is not translation invariant suggesting the bound on $\alpha$ and regret bound might be modified and improved under different extensions.
Nonetheless, the threshold is not merely an artifact of the proof. In simulations, given in Section \ref{Simulations}, we note that large values of $\alpha$ may yield a recurrent chain. From a practical stand point, this can be easily rectified by slowly decreasing $\alpha$. Alternatively, decreasing $\alpha$ every time $a_\star(t)$ changes will likely yield a logarithmic regret.

Notice the form of Theorem \ref{main_theorem} is interesting. The transient probability $Q$, which is commonly analysed in the stability theory of Markov chains, can be interpreted as the initial cost of exploration. Then after the last time $\{ p_{a^\star}(t) \leq 1/2\}$ holds, the Markov chain undergoes a transient stage where probability of the optimal arm converges to $1$ at rate $1/t$.

For Theorem \ref{3rd_theorem}, cooling schedules similar to \eqref{alpha_2nd_cond_III} are readily employed for the analysis of optimization in Markov chains, see \cite{bertsimas1993} and \cite{hajek}. This was the initial motivation for this learning parameter. 
The regret from the Lai and Robbins' lower-bound is known to be of order $\log T$ \cite{lai1985asymptotically}. There is a further multiplicative factor of $\log ( e+\log(T))$ in Theorem \ref{3rd_theorem}. However, we note that this function is very slowly increasing, e.g. $\log(e+\log(T))\leq 8$ for $T= 10^{1000}$ (achieving $T$ floating point operations is many orders of magnitude beyond modern computing power). So from a practical stand point this factor has negligible impact on the regret. This, along with Theorem~\ref{4th_theorem}, emphasizes the point that any disparity from $\log (T)$ regret can be controlled and made arbitrarily small by simple adjustments to the basic algorithm. In the case of SAMBA operating with averaged rewards, again we see dependence on the gap $\Delta$. However, since we now allow ourselves to estimate rewards, we can estimate $\Delta$. For example, if there is one optimal arm, we can let $\hat \Delta = \hat r_{a_\star} - \max_{a\neq a^\star} \hat r_a$ and then select $\alpha$ so that the conditions of Theorem \ref{ThrmAv} are satisfied. \footnote{Note, if there is more than one optimal arm then that $\hat \Delta = \hat r_{a_\star} - \max_{a\neq a^\star} \hat r_a$ goes to zero. However, by the Law of Iterated Logarithm, all optimal arms must be within an error $ \epsilon_a := \sqrt{\log \log t_a/t_a}$ of their mean. Thus we can take $\hat \Delta := \hat r_{a_\star} - \max_{a} \{  \hat r_a : \hat r_a + \epsilon_{a}  < \hat r_{a^\star } - \epsilon_{a^\star } \}$ instead.}

We assume Bernoulli distributed rewards. The assumption $\{0,1\}$ simplifies the proof in several ways. For instance, the process $(\bm p(t) : t\in \mathbb Z_+)$ is a countable state space Markov chain in this case. So we do not need to appeal to the more general theory of Harris chains \citep{meyn2012markov}. The results proven will extend to the case of bounded rewards, where the maximum reward is know. After normalizing rewards so that the maximum reward is $1$ and the minimum reward is $0$, the SAMBA algorithm can be implemented without change. 
The algorithm does not apply directly to unbounded rewards as we require the algorithm to maintain probabilities within the probability simplex. An unbounded reward could result in a transition outside the probability simplex. Other algorithms such a GBA deal with this feature by parameterizing the probability simplex. However, it remains an open problem to determine the regret of these algorithms in stochastic environments.

%% file: MODEL.tex
\section{Model and Notation.}\label{sec:model}

We describe a multi-armed bandit problem with a finite number of arms and Bernoulli distributed rewards. We also describe and give notation for the SAMBA process.

\smallskip
\noindent
\subsection{Arms and Rewards.} There is a finite set of arms $\mathcal A$ of cardinality $N:=|\mathcal A|$. At each time $t\in \mathbb Z_+$, you may choose an arm $a\in \mathcal A$. (Here $\mathbb Z_+ := \{ 0,1,2,...\}$.) When played, arm $a$ produces a reward that is a Bernoulli random variable. That is 
\[
R_a(t) = 
\begin{cases}
1 & \text{ w.p. }\, r_a\, , \\
0 & \text{ w.p. }\, 1- r_a\, ,
\end{cases}
\] 
where $r_a \in [0,1]$. (We use ``w.p.'' to abbreviate ``with probability''.) We assume that the random variables $(R_a(t): t\in \mathbb Z_+, a\in\mathcal A)$ are independent.  
We let the optimal arm and reward be
\[
a^\star = \argmax_{a\in \mathcal A}\, r_a,
\qquad 
r^\star = \max_{a\in \mathcal A}\, r_a\, .
\]
We assume the optimal arm is unique. The gap between the best arm and next best arm is
$
\Delta  : = r^\star - \max_{a: a\neq a^\star } r_a \, .
$

\subsection{Policies.} A policy chooses one arm to play at each time, and may use information of the past arms played and their rewards. More formally, 
we let $I_a(t)$ be the indicator function that arm $a$ is played at time $t$. We can summarize if arm $a$  was played at time $t$ and its reward with  $ (I_a(t), I_a(t) \cdot R_a(t) ) $. If $a$ is not played then  $ (I_a(t), I_a(t)\cdot R_a(t) )=(0,0) $ and if $a$ is played then $ (I_a(t), I_a(t)\cdot R_a(t) )=(1,R_a(t)) $. The history at time $t$ is then
$
H(t) := ((I_a(s), I_a(s)\cdot R_a(s)) : a\in \mathcal A, s\leq t).
$

A policy is any mechanism for choosing arms where, for each time $t$, the arm played $I_a(t)$, $a\in\mathcal A$, is a function of the history $H(t-1)$ and, perhaps, an independent uniform $[0,1]$ random variable used for randomization. 
Given we allow for randomization, it is useful to define
\[
p_a(t) := \mathbb E [ I_a(t) | H(t-1)],  
\]
{for} $a\in \mathcal A, t\in \mathbb Z_+$. Here $p_a(t)$ gives the probability of choosing $a$ at time $t$ given the past arms played and rewards received.
We let $\bm p(t) = (p_a(t) : a\in \mathcal A)$.
We define 
$
q_a(t) := 1-p_a(t).
$

\subsection{Regret.} The cumulative reward of a policy by time $T$ is
$$
\mathbb E \left[ \sum_{t=0}^{T-1} \sum_{a\in \mathcal A} I_a(t) R_a(t) \right].
$$
For example, note that the cumulative reward for playing the optimal arm at each time is $r^\star T$. This is the optimal policy; however, we focus on policies where the average reward for each arm is unknown.
The regret of a policy by time $T$, $\mathcal R\! g(T)$, is the expected difference between the cumulative reward from playing the best arm and the cumulative reward of the policy played. 
That is
\begin{equation}\label{Regret}
\mathcal R\! g(T)
: = 
r^\star T - \mathbb E \left[ \sum_{t=0}^{T-1} \sum_{a\in \mathcal A} I_a(t) R_a(t) \right]
=
\sum_{a: a\neq a^\star}
(r^\star - r_a) 
\mathbb E 
\left[
\sum_{t=0}^{T-1} p_a(t) 
\right]\,.
\end{equation}
The 2nd equality above is a straight-forward calculation, see Lemma \ref{Regret_Lemma}. It was shown by \citet{lai1985asymptotically} that $\mathcal R\! g(T)= \Omega (\log(T))$ provides a lower-bound for all asymptotically consistent policies. 

\subsection{SAMBA process.}
We now define notation for our policy. Here probabilities $(p_a(t) : a\in \mathcal A)$ are maintained for each arm. We let $a_\star (t)$ be the arm with maximal probability at time $t$ and we let $p_\star(t)$ be its probability, i.e.
\[
a_\star (t) = \argmax_{a\in \mathcal A} p_a(t)\, 
\qquad \text{and}\qquad 
p_{\star} (t) = \max_{a\in \mathcal A} p_a(t)\, .
\] 
If the maximum is not unique we select $a_\star(t)$ at random amongst the set of maximizing arms. 
We will often refer to $a_{\star}(t)$ as the \emph{leading arm}.
Further, we use the shorthand
\[
I_{\star}(t) = I_{a_\star(t)}(t),
\qquad 
R_{\star}(t) = R_{a_\star(t)}(t),\qquad 
r_{\star}(t) = \mathbb E [ R_\star(t) | H(t-1) ]\, .
\]
Note $R_{\star}(t)$ is \emph{not} the reward from the optimal arm, but the reward of the arm $a_{\star}(t)$ at time $t$.

We update the probabilities of each arm $a\neq a_\star(t)$ according to the rule
\begin{equation}
\label{update}
p_a(t+1) 
:=
p_a(t) 
+ 
\alpha 
p_a(t)^2 
\left[ 
\frac{I_a(t) R_a(t)}{p_a(t)}
-
\frac{I_{\star}(t) R_\star(t)}{p_\star(t)}
\right]	\, ,
\end{equation}
and, for the arm $a_\star(t)$, $p_{a_\star(t)}(t+1):= 1- \sum_{a\neq  a_{\star}(t)} p_a(t+1)$.
%
Given $(p_a(t) : a\in\mathcal A)$ has positive entries which sum to 1, the updated the vector $(p_a(t+1) : a\in \mathcal A)$ also has positive entries which also sums to 1. See Lemma \ref{lemma3} in Section \ref{app:lemma3} for a proof.


\subsection{Additional Notation.} 
We define additional notation, used later. We let $\wedge$ and $\vee$ denote the pairwise minimum and maximum, respectively, that is $x\wedge y = \min \{x,y \}$ and $x\vee y = \max \{x,y \}$.
For $a,b\in \mathbb Z$ we let $[a : b)= \{ a, a+1,..., b-1\}$.
We let $\mathcal P$ be the set of probability vectors on $\mathcal A$, that is
$
\mathcal P := \big\{ (p_a: a\in\mathcal A) : 
\sum_{a\in\mathcal A} p_a =1,
p_a \geq 0 \big\} \, .
$
We follow the convention that multiplication precedes division with a slash, e.g. $8/2x = 2$ for $x=2$. We apply the convention that $\log^n x = (\log x)^n$. 

%% file: THEOREM_1.tex
\section{Proof of Theorem \ref{main_theorem}.}
\label{sec:Proofs}

We organize the proof of Theorem \ref{main_theorem} into four parts, each a subsection below. Each subsection proves one main Proposition or Theorem. Some supporting lemmas and corollaries are proven in the Appendix.
First, in Section \ref{sec:recurrence}, we analyze the recurrence time of the chain to states with $q_{a^\star}  < {1}/{2}$.  
Specifically, in Corollary \ref{prop1}, we prove the expected recurrence time for this set is finite.
Second, in Section \ref{sec:embedding}, 
we define $\hat q(s)$ as the process that follows  $q_{a^\star}(t)$ when inside the set of states $\{ q_{a^\star} < \frac{1}{2} \}$. Proposition \ref{LemExpectq} shows that $\hat q(s)$ converges to zero and bounds its expected value.
Third, in Section \ref{sec:transience}, we analyse the transient behavior of our chain. We show in Proposition \ref{trans_prob_prop} that the event $q_{a^\star}>\frac{1}{2}$ is transient and that $\lim_{t\rightarrow\infty}q_{a^\star}(t)=0$. In other words the process converges on the optimal arm.
Fourth, in Section \ref{sec:regret},
we prove the regret bound required to complete Theorem \ref{main_theorem}.


\subsection{Recurrence Times.}
\label{sec:recurrence}

Lemma \ref{upperbound} is a discrete time analogue of the differential equation $\dot q(t) = - \eta q(t)^2$ that we analysed in Section \ref{Heuristic}. 

\begin{lemma}\label{upperbound}
	If $(\bar q(t) : t\in\mathbb Z_+) $ is a sequence of positive real numbers such that  
	\begin{equation}\label{qbound}
	\bar q(t+1) \leq \bar q(t) - \eta \bar q(t)^2	
	\end{equation}
	for some $\eta >0$, then, for all $t\in \mathbb Z_+$,
	\[
	\bar q(t) \leq \frac{\bar q(0)}{1+ \eta \bar q(0) t}\, .
	\]
\end{lemma}

\Beginproof
	Dividing the expression \eqref{qbound} by $\bar q(t)^2$ gives
	\begin{equation*}
	\frac{\bar q(t+1) - \bar q(t)}{\bar q(t)^2	} \leq - \eta \, .
	\end{equation*}
	Since $\bar q(t+1)$ is less than $\bar q(t)$ but still positive, dividing by $\bar q(t+1)$ rather than $\bar q(t)$ decreases the previous lower bound 
	\begin{equation*}
	\frac{\bar q(t+1) - \bar q(t)}{\bar q(t)	\bar q(t+1)} 
	= 
	\frac{1}{\bar q(t)} - \frac{1}{\bar q(t+1)} \leq - \eta \, .
	\end{equation*}
	Summing from $t=0,...,T-1$ gives 
	\[
	\frac{1}{\bar q(0)}-
	\frac{1}{\bar q(T)}
	\leq - \eta T\, ,
	\]
	which rearranges to give 
	\[
	\bar q(T) \leq \frac{\bar q(0)}{1+ \eta \bar q(0) T},
	\]
	as required.
\Endproof

We need to show that $p_{a^\star}$ does not get too small for too long. To make this precise, we define 
\begin{equation}
\label{TauE}
\tau( \hat x) :=
\min \Big\{ t \geq 1 : p_{a^\star}(t) > \frac{1}{\hat x} \Big\},
\qquad 
E(\hat x)
:=
\Big\{
\bm p \in \mathcal P :
\frac{1-\alpha}{\hat x}  \leq
p_{a^\star}(t) < \frac{1}{\hat x} 
\Big\}\, ,
\end{equation}
for $\hat x >0$.
Notice if $p_{a^\star}(0)\geq 1/\hat x$ and $p_{a^\star}(1) <  1/\hat x$ then $\bm p(1)\in E(\hat x)$. Proposition \ref{first_lemma} shows that, for sufficiently large $\hat x$, the expected time to return to $p_{a^\star}>\frac{1}{\hat x}$ is finite.

\begin{proposition}\label{first_lemma}
	For $\alpha>0$ such that \eqref{alpha_cond} holds,
	there exists a positive constant $\hat x$ such that 
	\[
	\sup_{\bm p \in E(\hat x)}
	\mathbb E \big[ \tau(\hat x) 
	|
	\bm p(0) = \bm p
	\big]
	< \infty \, .
	\]
\end{proposition}
\Beginproof
	We will show that $
	{1}/{p_{a^\star}(t)} - c t$
	is a supermartingale for some $c >0$. The proof will then follow by the Optional Stopping Theorem. 
	
	First we collect together some constants to define $\hat x$. 
	Notice if \eqref{alpha_cond} holds then $1+\alpha < r^\star / (r^\star - \Delta)$.
	Thus there exists an $\epsilon>0$ such that
	$
	(1+\epsilon)(1+\alpha) < {r^\star}/{(r^\star - \Delta)}
	$
	or, equivalently,
	\[
	c:=
	\alpha \frac{r^\star}{1+\alpha}
	-
	\alpha (r^\star-\Delta)(1+\epsilon) >0\, .
	\]
	Given this choice of $\epsilon$, we define $\hat x$ to be greater than $N$ and such that 
	\begin{equation}\label{lem:xbound}
	\frac{\hat x}{\hat x - \alpha N}
	\leq 1+\epsilon\, .	
	\end{equation}
	
	We assume that $p_{a^\star}(0) \in E(\hat x)$.
	Note that, for $t=1,...,\tau(\hat x)-1$ , $a^\star$ is not the leading arm and so $p_{a^\star}(t)$ obeys the update equation \eqref{update}. That is
	$p_{a^\star} (t+1) =(1+\alpha) p_{a^\star} (t)$ 
	with probability $p_{a^\star}(t) r^\star$ and $p_{a^\star} (t+1)= p_{\star}(t) r_\star(t)$ with probability $p_{\star}(t) r_\star(t)$.
	So, if we let $x(t) = p_{a^\star}(t)^{-1}$ then, a short calculation gives, 
	\[
	x(t+1) 
	=
	\begin{cases}
	x(t) 
	-
	\frac{\alpha}{1+\alpha}
	x(t)
	&
	\text{ w.p. } 
	\frac{r^\star}{x(t)}\, ,
	\\
	x(t)
	+
	\alpha 
	\frac{	
		x (t)
	}{
		p_\star (t) x(t) - \alpha  
	}
	& \text{ w.p. } p_{\star}(t) r_{\star}(t)\, ,
	\\
	x(t) & \text{ otherwise.}
	\end{cases}
	\]
	Thus
	\begin{align*}
	\mathbb E [
	x(t+1) 
	|
	H(t)
	]
	- x(t)
	&
	=
	\alpha
	r_{\star} (t) 
	\frac{p_{\star}(t)x(t)}{
		p_{\star}(t) x(t) - \alpha
	}
	-
	\frac{\alpha r^\star}{1+\alpha} 
	\leq 
	\alpha (r^\star - \Delta)
	(1+\epsilon) 
	-
	\frac{\alpha r^\star}{1+\alpha} 
	= -c.
	\end{align*}
	The above inequality holds by $r_{\star}(t) \leq r^\star - \Delta $ 
	and by \eqref{lem:xbound}
	[and noting $p_{\star}(t)> 1/N$].
	The constant $c$ is as given above.
	
	By the Optional Stopping Theorem,
	\begin{equation}
	\label{xrbound}
	- c \mathbb E [\tau(\hat x) \wedge t]
	\geq 
	\mathbb E [ x(\tau(\hat x) \wedge t)]
	-
	\mathbb E [ x(0) ] 
	\geq 
	-
	\frac{\hat x}{1-\alpha}\, .
	\end{equation}
	The final inequality holds since
	$x(\tau(\hat x) \wedge t)$ is positive and  since
	$
	x(0) = 1/p_{a^\star}(0) \leq \hat x / (1-\alpha)$ holds when $\bm p(0) \in E(\hat x)$. 
	Finally applying the Monotone Convergence Theorem to \eqref{xrbound} gives that
	$
	\mathbb E [ \tau(\hat x) ]
	\leq {\hat x}/{c(1-\alpha)}
	$
	as required.    
	\Endproof

 The result below extends Proposition \ref{first_lemma} to a hitting times for a slightly smaller region, where $\hat x = 2$. Note this bounds the expected time that it takes for the optimal arm to return to being the leading arm.

\begin{corollary}
	\label{prop1}
	For $\alpha>0$ such that \eqref{alpha_cond} holds,
	there exists a positive constant $\hat x$ such that 
	\[
	\sup_{\bm p \in E(2)}
	\mathbb E \big[ \tau(2) 
	|
	\bm p(0) = \bm p
	\big]
	< \infty \, .
	\]	
\end{corollary}
Corollary \ref{prop1}, proven in Section \ref{corol5},  follows from standard Markov chain arguments. We show that the probability of reaching $p_{a^\star} > \frac{1}{2}$ before reaching $p_{a^\star} \leq \frac{1}{\hat x}$ is bounded below. (For instance, this will occur from a long run of rewards from arm $a^\star$.) Since the return time to $p_{a^\star} > \frac{1}{\hat x}$ is finite, there are a geometric number of finite expectation trials to reach $p_{a^\star} > \frac{1}{2}$.

%

\subsection{An Embedded Chain.}
\label{sec:embedding}

We know that $p_{a^\star} > \frac{1}{2}$, or equivalently $q_{a^\star}< \frac{1}{2}$, must occur in finite time. 
We now analyse the process $q_{a^\star}(t)$ for values less than $1/2$ for this reason we define the embedded process $\hat q (s)$, $s\in\mathbb Z_+$ which we now describe.

\begin{figure}
	\includegraphics[width=\textwidth]{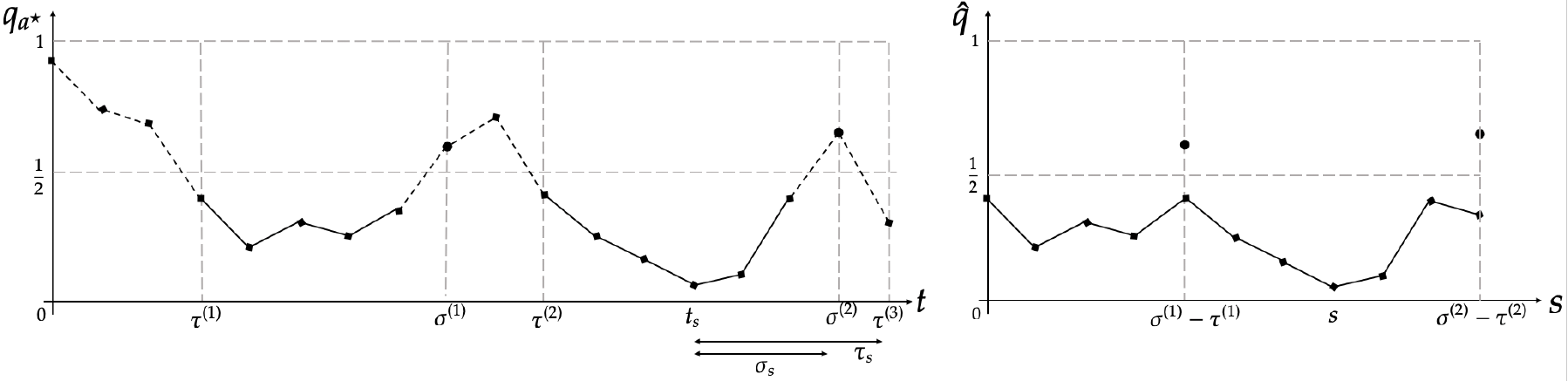}
	\caption{Here $\hat q(s)$ is constructed from $q_{a^\star}(t)$ by deleting areas between the vertical grey lines. Also, above are examples of stopping times $\tau^{(k)}$, $\sigma^{(k)}$, $s$, $t$, $\sigma_s$ and $\tau_s$.\label{Fig1}}
\end{figure}

We define the following stopping times: 
$\tau^{(0)} = 0$,
$\sigma^{(0)} = 0$,
and, for $k\in \mathbb N$,
\begin{align}
\sigma^{(k)}
=
\min 
\left\{
t \geq \tau^{(k)}
:
q_{a^\star} (t) 
\geq \frac{1}{2}
\right\},	
\qquad
\tau^{(k)}
=
\min 
\left\{
t \geq \sigma^{(k-1)}
:
q_{a^\star} (t) 
< \frac{1}{2}
\right\}\, .
\label{tau_stopping}
\end{align}
If no such time exists for some $k$ in \eqref{tau_stopping} then we let $\sigma^{(k)}=\infty$ (resp., $\tau^{(k)}=\infty$ ). Notice the times in \eqref{tau_stopping} partition the times $\mathbb Z_+$ into regions $[\tau^{(k)} : \sigma^{(k)} )$ 
and
$[\sigma^{(k)} : \tau^{(k+1)} )$, $k\in\mathbb Z$ where 
\begin{align*}
q_{a^\star}(t)
<
\frac{1}{2},
\quad 
\text{for}
\quad 
t \in 
[\tau^{(k)} : \sigma^{(k)} ),\quad
\text{and}\quad 
q_{a^\star}(t)
\geq 
\frac{1}{2},
\quad 
\text{for}
\quad 
t \in 
[\sigma^{(k)} : \tau^{(k+1)} ).
\end{align*}
See Figure \ref{Fig1} for a representation of these times. 

We analyze the rate at which $p_{a^\star}(t)$ approaches one, or equivalently, the rate at which $q_{a^\star}(t)$ approaches zero. 
We consider the process that follows $q_{a^\star}(t)$ over intervals where $q_{a^\star}(t) < \frac{1}{2}$ but ignores times where $q_{a^\star}(t) \geq \frac{1}{2}$. Specifically, we define $( \hat{q}(s) : s\in \mathbb Z_+)$ by
\begin{align*}
\hat q(s) 
:=
q_{a^\star} 
(
t_s
) 
\qquad 
&\text{where}\qquad 
t_s=
s 
+ 
\sum_{i=0}^k 
\big( 
\tau^{(i+1)} 
-
\sigma^{(i)}
\big)
\\
&
\text{for}
\qquad\quad 
s\in 
\bigg[
\sum_{i=0}^k
(\sigma^{(i)} 
-\tau^{(i)} )
:
\sum_{i=0}^{k+1}
(\sigma^{(i)} 
-\tau^{(i)} )
\bigg),\quad	k\in \mathbb Z_+.
\end{align*}
Again see Figure \ref{Fig1} for a more intuitive representation of $\hat q(s)$.
Further we let
\[
\sigma_s
=
\min \Big\{
t>0 
:
p_{a^\star}(t+t_s) > \frac{1}{2}
\Big\}\, ,
\quad
\text{and}
\quad
\tau_s
=
\min \Big\{
t> \sigma_s
:
p_{a^\star}(t+t_s) \leq \frac{1}{2}
\Big\}\, .
\]
See Figure \ref{Fig1}, which plots instances of $s$, $t_s$, $\sigma_s$ and $\tau_s$.

Proposition \ref{LemExpectq} shows that $\mathbb E [ \hat q(s) ] = \mathcal O ( \frac{1}{s})$.

\begin{proposition}
	\label{LemExpectq}
	a) The process $(\hat q(s) : s\in \mathbb Z_+)$ is a positive supermartingale
	\begin{equation}\label{qIneq}
	\mathbb E [
	\hat q ( s+1) 
	| \hat H(s) ]
	-
	\hat q(s) 
	\leq 
	- \frac{\alpha \Delta}{N  }
	\hat q(s)^2\, .
	\end{equation}
	b) With probability $1$,
	$
	\hat q(s) \xrightarrow[]{} 0\, ,
	$ as $s\rightarrow \infty$ .\\
	c) 
	\begin{equation}\label{qTime}
	\mathbb E [ \hat q (s) ]
	\leq 
	\frac{ N}{
		2N 
		+
		\alpha \Delta s 
	}\, .	
	\end{equation}
\end{proposition}
\Beginproof
Note that $t_s$, $\sigma_s$ and $\tau_s$ are stopping times with respect to the history $H(t)$. So the Strong Markov Property applies to the process $(\bm p(t) : t\in \mathbb Z_+)$ at these times.
Since $t_s$ is a stopping time for each $s\in \mathbb Z_+$, $\hat q(s)$ is adapted to history $\hat H (s) := H(t_s)$.

	a) Observe that by definition $\hat q(s) = q_{a^\star}(t_s)$ and that 
	\begin{equation*}
	\hat q(s+1) 
	=
	\begin{cases}
	q_{a^\star} (t_s+1) 
	&
	\text{if }
	q_{a^\star}(t_s+1) < \frac{1}{2}\, , 
	\\
	q_{a^\star} (t_s + \tau_s )
	&\text{if } 
	q_{a^\star}(t_s+1) \geq \frac{1}{2}\, . 
	\end{cases}
	\end{equation*}
	Since $q_{a^\star}(t_s+\tau_s) < \frac{1}{2}$, we have that 
	\begin{equation}
	\label{q1}
	\hat q (s+1)
	\leq 
	q_{a^\star} (t_s+1) \, .
	\end{equation}
	We know at times $t_s$, $a^\star$ is the leading arm. Thus, for all $a\neq a^\star$,
	\begin{equation*}
	p_a(t_s +1)
	=
	p_a(t_s) 
	+
	\alpha 
	p_a(t_s)^2 
	\left[
	\frac{I_a(t_s) R_a(t_s)}{p_a(t_s)}
	-
	\frac{I_{a^\star}(t_s) R_{a^\star}(t_s)}{p_{a^\star}(t_s)}
	\right]\, .
	\end{equation*}
	So taking expectations,
	\[
	\mathbb E [ p_a(t_s+1) -p_a(t_s) | H(t_s) ]
	=
	\alpha 
	p_a(t_s)^2 ( r_a -r_{a^\star})
	\leq 
	- \alpha \Delta p_a(t_s)^2\,.
	\]
	Since $q_{a^\star}$ is defined to be the sum of $p_a$ over $a\neq a^\star$, we have that
	\begin{equation}\label{q3}
	\mathbb E [ q_{a^\star}(t_s+1) -q_{a^\star}(t_s) | H(t_s) ]
	\leq -\Delta \sum_{a: a\neq a^\star} 
	\alpha  p_a(t_s)^2
	\, .
	\end{equation}
	Recalling that $\hat H(s) := H(t_s)$ and combining \eqref{q1} and \eqref{q3} gives 
	\begin{align}
	\mathbb E [ \hat q(s+1) | \hat H(s)  ]
	-
	\hat q(s) 
	&
	\leq 
	\mathbb E [
	q_{a^\star} (t_s+1)
	-
	q_{a^\star}(t_s)
	| H(t_s) ]
	\leq 
	- \Delta \sum_{a: a\neq a^\star} 
	\alpha p_a(t_s)^2\, .
	\label{q4}
	\end{align}
	By Jensen's inequality:
	\begin{align*}
	\sum_{a: a\neq a^\star} 
	p_a(t_s)^2
	&
	=
	(N-1) 
	\sum_{a: a\neq a^\star} 
	\frac{p_a(t_s)^2}{N-1}
	\geq 
	(N-1)
	\bigg(
	\sum_{a: a\neq a^\star} 
	\frac{p_a(t_s)}{N-1}
	\bigg)^2
	=
	\frac{
		q_{a^\star}(t_s)^2}{
		N-1}
	\geq 
	\frac{	
		\hat q(s)^2}{N}\, .
	\end{align*}
	Thus applying the above bound to \eqref{q4} gives 
	\[
	\mathbb E [ \hat q(s+1) | \hat H(s) ]
	-
	\hat q(s) 
	\leq 
	-\frac{\alpha \Delta }{N}
	\hat q(s)^2 \, ,
	\]
	which is the required bound. 
	It is immediate, from the above bound that $\hat q(s)$ is a supermartingale. \smallskip
	
	\noindent b) By definition $\hat q(s)$ is positive. So by Doob's Supermartingale Convergence Theorem the limit $\lim_{s\rightarrow\infty} \hat q(s)$ exists. We now show that this limit is zero.
	
	Since the limit exists it is sufficient to show that $\liminf_s \hat q(s) =0 $. For $m > 2$, let 
	\[
	\phi_m = \min \Big\{ s \geq 1 : \hat q(s) < \frac{1}{m} \Big\}.
	\]
	It is sufficient to show that $\phi_m<\infty$, with probability $1$, since then it is clear that we can define a sequence of stopping times 
	$
	\psi_m := \min \{ s \geq  \psi_{m-1} : \hat q(s) < \frac{1}{m} \}
	$
	each of which is finite, w.p. 1, and $\hat q (\psi_m) \rightarrow 0$, which implies $\liminf_{s\rightarrow 0} \hat q(s) =0
	$.
	
	To show that $\phi_m$ is finite, notice that by part a),
	\[
	\mathbb E [ \hat q(s+1) | \hat H(s) ]
	-
	\hat q(s) 
	\leq 
	-\frac{\alpha}{N}
	\Delta 
	\hat q(s)^2 
	\leq
	-\frac{\alpha}{N}
	\Delta 
	\frac{1}{m^2}
	\]
	where the last inequality holds for so long as $\hat q (s) \geq \frac{1}{m}$. Thus by the Optional Stopping Theorem
	\[
	\mathbb E [ \hat q( \phi_m \wedge s) ]
	-
	\mathbb E [ \hat q(0) ]
	\leq 
	-
	\frac{\alpha}{N} \Delta \frac{1}{m^2} \mathbb E [ \phi_m \wedge s  ]\, .
	\]
	Rearranging and applying Monotone Convergence Theorem gives
	\[
	\mathbb E [ \phi_m ]
	\leq 
	\lim_{s\rightarrow \infty } \mathbb E [\phi_m\wedge s  ]  
	\leq \frac{Nm^2}{\alpha \Delta} \mathbb E[ \hat q(0) ] < \infty .
	\]
	Thus with probability $1$, $\phi_m <\infty$ as required. Thus, as show above, $\liminf_{s\rightarrow \infty} \hat q(s) = 0$ and so $\lim_s \hat q(s)=0$, as required.\smallskip
	
	\noindent c) Finally, the required bound \eqref{qTime} holds by taking expectations in part a) and applying Lemma \ref{upperbound}:
	\begin{align*}
	\mathbb E [\hat q(s) ] 
	\leq 
	\frac{
		\mathbb E [ \hat q(0) ]
	}{ 
		1
		+
		\frac{\alpha}{N} \Delta \mathbb E [ \hat q(0) ] s
	}
	\leq 
	\frac{N}{2N + \alpha \Delta s }\, .
	\end{align*}
	Above, we use that $x\mapsto x/(1+x)$ is increasing for $x>0$ and $\mathbb E [ \hat q(0) ] < 1/2 $.
   
\Endproof

\subsection{Transience.}
\label{sec:transience}

With Corollary \ref{prop1}, we know that $q_{a^\star} \leq \frac{1}{2}$ occurs in finite time and, by Proposition \ref{LemExpectq}, we know that $q_{a^\star}$ goes to zero. We combine these two results to prove the transience of $(\bm p(t) : t\in\mathbb Z_+)$. 
Proposition \ref{trans_prob_prop} below collects together these results. 

\begin{proposition}
	\label{trans_prob_prop}
	a) If $q_{a^\star}(0) \leq  \frac{1}{2}$ then, there exists a constant $\rho < 1$ such that 
	$
	\mathbb P ( \sigma^{(1)} < \infty ) < \rho\, .		
	$
	Moreover
	\begin{equation*}
	\mathbb P ( \sigma^{(k)} < \infty | \sigma^{(k-1)} <\infty)
	< \rho \, .
	\end{equation*}
	b) 
	\[
	\sum_{t=0}^\infty
	\mathbb P 
	\Big(
	q_{a^\star} (t) \geq \frac{1}{2}
	\Big)	
	< \infty \, .
	\]
	c) With probability $1$,
	$
	q_{a^\star}(t) \rightarrow
	0$ as $t\rightarrow \infty$.
\end{proposition}

\Beginproof
	a)
	We know that at time $0$ there is a probability greater than a half of trying arm $a^\star$ and a positive probability of receiving a reward greater than $r^\star$.
	On this event
	if $1/4 <q_{a^\star}(0) \leq {1}/{2}$, we know that 
	\[
	q_{a^\star}(1)
	\leq 
	q_{a^\star}(0) 
	- 
	\frac{\alpha}{N} r^* q_{a^\star}(0)^2
	\leq \frac{1}{2} - \frac{\alpha}{16 N} < \frac{1}{2} \, ,
	\] 
	and if $q_{a^\star}(0) \leq \frac{1}{4}$ then 
	\[
	q_{a^\star}(1)
	\leq 
	q_{a^\star} + \alpha q_{a^\star}(0)
	\leq 
	\frac{1+\alpha}{4}
	< \frac{1}{2}.
	\]
	Thus there is a positive probability, say $p'$, that if $q_{a^\star}(0) \leq \frac{1}{2} $ then at the next time 
	$q_{a^\star}(1) \leq c < 1/2$ for $c= ({1}/{2} - \alpha/16 N)\vee ((1+\alpha)/4)$.
	
	By Proposition \ref{LemExpectq}, $q_{a^\star}(t\wedge \sigma^{(1)})$ is a supermartingale and if $\sigma^{(1)} = \infty$ then $q_{a^\star}(t) = \hat q (t) \rightarrow 0$ as $t\rightarrow\infty$.
		Now suppose that $q_{a^\star}(1) < c$, by the Optional Stopping Theorem,
	\begin{align*}
	c >
	\mathbb E [ q_{a^\star} (1) | q_{a^\star}(1) < c ]
	&\geq 
	\mathbb E[ q_{a^\star}(\sigma^{(1)} \wedge t ) | q_{a^\star}(1) < c  ] 
	\\
	&
	=
	\mathbb E [ 
	q_{a^\star} ( \sigma^{(1)} ) 
	\mathbb I [ \sigma^{(1)} \leq t ] | q_{a^\star}(1) < c]	
	+
	\mathbb E [ 
	q_{a^\star} ( t ) 
	\mathbb I [ \sigma^{(1)} > t ] | q_{a^\star}(1) < c]	
	\\
	&
	\geq 
	\frac{1}{2}
	\mathbb P ( \sigma^{(1)} \leq t | q_{a^\star}(1) < c ) 
	+
	\mathbb E [
	\hat q (t) \mathbb I [ \sigma^{(1)} > t ]
	| q_{a^\star}(1) < c
	]	
	\\
	&
	\xrightarrow[t\rightarrow \infty	]{}
	\frac{1}{2} \mathbb P( \sigma^{(1)} < \infty | q_{a^\star}(1) < c) \, .
	\end{align*}
	In the final limit we apply the Dominated Convergence Theorem and the fact that $\hat q (t) \rightarrow 0$ with probability $1$.
	Thus
	\[
	\mathbb P ( \sigma^{(1)} = \infty | q_{a^\star}(1) < c  ) > (1-2c) =: \delta .
	\]
	The above holds given at time $1$, $q_{a^\star}(1) < c$. However, as discussed above, there is a positive probability, say $p'$, of reaching state $q_{a^\star}(1) < c$. Thus
	\[
	\mathbb P\Big(\sigma^{(1)} = \infty \Big| q_{a^\star}(0) \leq \frac{1}{2}\Big)
	\geq 
	(1-p')
	+p' \mathbb P ( \sigma^{(1)} = \infty | q_{a^\star} (1) < c)
	\leq 1- (1-\delta ) p 
	=: \rho
	\, .
	\]
	This gives the first bound required in part a). 
	
	For the second bound, by the Markov property we know that 
	\[
	\mathbb P(
	\sigma^{(k)}
	< \infty 
	|
	\tau^{(k-1)} 
	< \infty
	)
	< \rho \, .
	\]
	Note that 
	\[
	\{
	\sigma^{(k-1)}<\infty 
	\}
	\cap 
	\{
	\tau^{(k-1)} 
	-
	\sigma^{(k-1)}
	<\infty 
	\}
	=
	\{
	\tau^{(k-1)}<\infty 
	\}
	\]
	where 
	$\mathbb P ( \tau^{(k-1)} - \sigma^{(k-1)} < \infty 
	|
	\sigma^{(k-1)} <\infty
	) = 1 $,
	by Corollary \ref{prop1}.  So $\mathbb P(	\sigma^{(k-1)} 
	< \infty
	) = \mathbb P(	\tau^{(k-1)} 
	< \infty
	)$ .
	Thus we have that
	\begin{equation*}	\mathbb P (
	\sigma^{(k)} < \infty
	|
	\sigma^{(k-1)} 
	< \infty
	)
	=
	\mathbb P(
	\sigma^{(k)}
	< \infty 
	|
	\tau^{(k-1)} 
	< \infty
	)
	< \rho \, ,
	\end{equation*}
	for $\rho<1$. 
	
	\noindent b) We know $q_{a^\star}(t) \geq \frac{1}{2}$ holds only for the time intervals 
	$[\sigma^{(k)} : \tau^{(k+1)} )$. So we have
	\begin{align}
	\sum_{t=0}^\infty
	\mathbb P 
	\Big(
	q_{a^\star} (t) \geq \frac{1}{2}
	\Big)	
	&
	=
	\mathbb E
	\left[
	\sum_{t=0}^\infty
	\mathbb I
	\Big[
	q_{a^\star}(t)
	\geq 
	\frac{1}{2}
	\Big]
	\right]
	=
	\mathbb E
	\left[
	\sum_{k=0}^\infty 
	\Big(
	\tau^{(k+1)}
	-
	\sigma^{(k)}
	\Big)
	\mathbb I
	\Big[
	\sigma^{(k)}
	<
	\infty
	\Big]
	\right]
	\notag
	\\
	&
	=
	\sum_{k=0}^\infty 
	\mathbb E 
	\left[
	\mathbb E 
	\big[
	\tau^{(k+1)}
	-
	\sigma^{(k)}
	\big|
	H(\sigma^{(k)})
	\big]
	\mathbb I
	\big[
	\sigma^{(k)}
	<
	\infty
	\big]
	\right]\, .
	\label{HTag}
	\end{align}
	We analyze the terms in the summands of \eqref{HTag}. 
	By Corollary \ref{prop1} and the Strong Markov Property, we have that, for a finite constant $C$,
	\begin{equation}\label{expected_return}
	\mathbb E 
	\big[
	\tau^{(k+1)}
	-
	\sigma^{(k)}
	\big|
	H(\sigma^{(k)})
	\big]
	< {C}	
	\end{equation}
	on the event $\{ \sigma^{(k)} < \infty \}$.

	So applying \eqref{expected_return} to the summand in \eqref{HTag} and applying part a), we have that 
	\begin{align*}
	\mathbb E 
	\left[
	\mathbb E 
	\big[
	\tau^{(k+1)}
	-
	\sigma^{(k)}
	\big|
	H(\sigma^{(k)})
	\big]
	\mathbb I
	\big[
	\sigma^{(k)}
	<
	\infty
	\big]
	\right]
	&
	\leq 
	C \mathbb P( \sigma^{(k)} < \infty ) 
	\\
	&
	=
	C
	\prod_{\kappa =1}^k
	\mathbb P( \sigma^{(\kappa )} < \infty | \sigma^{(\kappa -1)} < \infty ) = C\rho^k\, .
	\end{align*}
	Now applying this bound to \eqref{HTag}, gives the required bound,
	\[
	\sum_{t=0}^\infty 
	\mathbb P \Big( q_{a^\star} (t) > \frac{1}{2} \Big)
	\leq 
	\sum_{k=0}^\infty 
	C \rho^k 
	=
	\frac{C}{1-\rho} <\infty \, .
	\]
	c) By part b) , 
	$
	\mathbb E
	\big[
	\sum_{t=0}^\infty
	\mathbb I
	\big[
	q_{a^\star}(t)
	\geq 
	\frac{1}{2}
	\big]
	\big] < \infty \, .
	$ 
	Thus, with probability $1$, $
	\sum_{t=0}^\infty
	\mathbb I
	\big[
	q_{a^\star}(t)
	\geq 
	\frac{1}{2}
	\big] < \infty \, .$
	This implies that eventually $q_{a^\star}(t) < \frac{1}{2}$.
	From this time onwards the processes $q_{a^\star}$ and $\hat q$ are identical and we know that $\hat q(s) \rightarrow 0$.
	Specifically there exists a number $k_{\max} < \infty$ such that 
	$\tau^{(k_{\max})} <\infty$ but $\sigma^{(k_{\max})} = \infty$. Under the coupling above, 
	\[
	q_{a^\star}(t + \tau^{(k_{\max})})
	=
	\hat q
	\bigg(t 
	-
	\sum_{k=0}^{k_{\max}-1}
	(\tau^{(k+1)} -\sigma^{(k)})
	\bigg)
	\xrightarrow[t \rightarrow \infty]{} 0\, ,
	\]
	where the equality above holds for all $t\in \mathbb Z_+$ and the limit above holds by Proposition \ref{LemExpectq}b). Thus $\lim_{t\rightarrow\infty} q_{a^\star}(t) =0$ as required.

\Endproof 

\subsection{Rate of Convergence and Regret Bound.}
\label{sec:regret}

We prove the regret bound \eqref{regret_bound}. With this, we will have completed the proof of Theorem \ref{main_theorem}.

\begin{proof}[Proof of Theorem \ref{main_theorem}.]
	Note that by Proposition \ref{trans_prob_prop}c), we have that
	$
	p_{a^\star}(t) = 1- q_{a^\star}(t) \xrightarrow[]{} 1,
	$ as $t\rightarrow\infty$.
	So it remains to show the regret bound.
	Since $r^\star - r_a \leq 1$,
	\begin{equation}
	\label{Regret_q}
	\mathcal R\! g(T)
	=
	\sum_{a: a\neq a^\star}
	(r^\star - r_a)
	\mathbb E\left[
	\sum_{t=0}^{T-1}
	p_a(t)
	\right]
	\leq 
	\mathbb E 
	\left[
	\sum_{t=0}^{T-1}
	\sum_{a : a\neq a^\star}
	p_a(t)
	\right]
	=
	\mathbb E
	\left[
	\sum_{t=0}^{T-1}
	q_{a^\star}(t)
	\right]	.
	\end{equation}
	See Lemma \ref{Regret_Lemma} for the 1st equality above.
	We focus on bounding the final term in \eqref{Regret_q}.
	\begin{align}
	\mathbb E 
	\left[
	\sum_{t=0}^{T-1}
	q_{a^\star}(t) 
	\right]
	& =
	\sum_{t=0}^{T-1}
	\mathbb E 
	\left[
	q_{a^\star}(t)
	\mathbb I
	\Big[
	q_{a^\star}(t) 
	\geq
	\frac{1}{2}
	\Big]
	\right]
	+
	\mathbb E 
	\left[
	\sum_{t=0}^{T-1}
	q_{a^\star}(t)
	\mathbb I
	\Big[
	q_{a^\star}(t) 
	<
	\frac{1}{2}
	\Big]
	\right]
	\notag
	\\
	&
	\leq 
	Q
	+
	\mathbb E 
	\left[
	\sum_{t=0}^{T-1}
	q_{a^\star}(t)
	\mathbb I
	\Big[
	q_{a^\star}(t) 
	<
	\frac{1}{2}
	\Big]
	\right]
	\label{Qbound}
	\end{align}
	where, by Proposition \ref{trans_prob_prop},
	$
	Q := \sum_{t=0}^\infty 
	\mathbb P
	\Big(
	q_{a^\star}(t) 
	> 
	\frac{1}{2}
	\Big)
	< \infty .
	$
	
	We must show that the remaining term in \eqref{Qbound}
	grows logarithmically in $T$. Recalling the definition of $\hat q(s)$ from Section \ref{sec:embedding} and Figure \ref{Fig1}, for each $t$ such that $q_{a^\star}(t)<\frac{1}{2}$, there exists a corresponding value of $s$ with $s\leq t$ such that $\hat q(s) = q_{a^\star}(t)$. This gives the first inequality in the following sequence of inequalities, 
	\begin{align}
	\mathbb E 
	\left[
	\sum_{t=0}^{T-1}
	q_{a^\star}(t)
	\mathbb I
	\Big[
	q_{a^\star}(t) 
	<
	\frac{1}{2}
	\Big]
	\right]
	&
	\leq 
	\mathbb E \bigg[ 
	\sum_{s=0}^{T-1} 
	\hat q(s) 
	\bigg]
	\leq 
	\sum_{s=0}^{T-1} 
	\frac{N}{2N + \alpha \Delta s
	}
	\leq 
	\frac{N}{\alpha\Delta} 
	\log T\, .
	\label{half_Ineqs}
	\end{align}
	In the second inequality above, we apply Proposition \ref{LemExpectq}c). In the third inequality above we apply Lemma \ref{LemABC}.
	Applying \eqref{half_Ineqs} to \eqref{Qbound} and \eqref{Regret_q} gives the required bound
	\begin{align*}
	\mathcal R\! g(T)
	\leq \frac{N}{\alpha\Delta} \log T + Q\, .
	\end{align*}
\end{proof}

%% file: THEOREM_av.tex
\section{Proof of Theorem \ref{ThrmAv}.}
We now proceed with the proof of Theorem \ref{ThrmAv}. 
The proof is organized as follows. We provide some simple upper- and lower-bound on the SAMBA recursion (Lemma \ref{lem1} and Lemma \ref{lem2}). We then collect some standard concentration inequalities (Bernstein's Inequality, Lemma \ref{lem3} and a Chernoff Bound, Lemma \ref{lem4}). We then bound the empirical mean reward for each arm in terms of the number of iterations of the algorithm, Proposition \ref{prop1}. This part of the proof is analogous to the analysis of $\epsilon$-Greedy with decaying $\epsilon$, see \cite[Theorem 3]{auer2002finite}.  We use this to prove that there is a finite time where the optimal arm must have the best average reward by some margin, Proposition \ref{prop2}. We then apply this along with the upper-bound on the SAMBA recursion to prove the result. 

\begin{lemma}\label{lem1}
	If $p(t), t\geq 0$ is a sequence of numbers in the interval $[0,1]$ such that
\begin{align*}
  p(t+1) \geq p(t) - \alpha p(t)^2 , \qquad  t \geq 0,
\end{align*}
with $\alpha \leq 1/2$ then
\begin{align*}
  p(t) \geq \frac{1}{2 \alpha t + p(0)^{-1}}, \qquad t \geq 0 \, .
\end{align*}
\end{lemma}

\Beginproof
  We prove the result by induction. In particular suppose that at time $t$
\begin{align}\label{lem1:eq1}
  p(t) \geq \frac{c}{t+t_0}
\end{align}
where $t_0$, $c$ are constants that we will determine shortly. 
Given this, notice that 
\begin{align*}
  p(t+1) \geq p(t) - \alpha p(t)^2
\geq \frac{c}{t + t_0} 
-
\alpha 
\frac{c^2}{(t+t_0)^2}
\end{align*}
where the 2nd inequality holds since $p \mapsto p-\alpha p^2$ is an increasing function for $\alpha \leq 1/2$. 
Notice that given the above bound, the condition \eqref{lem1:eq1} holds at time $t+1$ provided
\begin{align*}
  \frac{c}{t+t_0} - \alpha \frac{c^2}{(t+t_0)^2}
\geq 
\frac{c}{t+1+t_0} \, .
\end{align*}
Rearranging shows that this is equivalent to the condition 
\begin{align} \label{lem1:eq2}
  t+t_0 \geq \frac{\alpha c}{1- \alpha c} \, .
\end{align}
In particular, if we take 
\begin{align}\label{lem1:ct0}
  c = \frac{1}{2 \alpha} \quad \text{and} \quad t_0 = \frac{p(0)^{-1}}{2\alpha}
\end{align}
then notice that the condition \eqref{lem1:eq1} holds at time $t=0$ and for any $t$ if \eqref{lem1:eq1} holds at time $t$ then it also holds at time $t+1$ (because \eqref{lem1:eq2} is satisfied). Thus the induction steps holds and, substituting \eqref{lem1:ct0} into \eqref{lem1:eq1}, we have 
\begin{align*}
  p(t) \geq \frac{1}{2 \alpha t + p(0)^{-1}}, \qquad t \geq 0 \, 
\end{align*}
as required. 
\Endproof

\begin{lemma}\label{lem2}
If $p(t), t \geq0$ is a sequence of numberrs in the interval $[0,1]$ such that 
\begin{align*}
  p(t+1) \leq p(t) - \gamma p(t)^2
\end{align*}
then 
\begin{align*}
  p(t) \leq \frac{1}{1+\gamma t}\, .
\end{align*}
\end{lemma}

\Beginproof
		Dividing the expression $p(t+1) \leq p(t) - \gamma p(t)^2$ by $p(t)^2$ gives
	\begin{equation*}
	\frac{p(t+1) - p(t)}{p(t)^2	} \leq - \eta \, .
	\end{equation*}
	Since $p(t+1)$ is less than $p(t)$ but still positive, dividing by $p(t+1)$ rather than $p(t)$ decreases the previous lower bound 
	\begin{equation*}
	\frac{p(t+1) - p(t)}{p(t)	p(t+1)} 
	= 
	\frac{1}{p(t)} - \frac{1}{p(t+1)} \leq - \eta \, .
	\end{equation*}
	Summing from $t=0,...,T-1$ gives 
	\[
	\frac{1}{p(0)}-
	\frac{1}{p(T)}
	\leq - \eta T\, ,
	\]
	which rearranges to give 
	\[
	p(T) \leq \frac{p(0)}{1+ \eta p(0) T} \leq \frac{1}{1+ \eta T}
	\]
	as required.
\Endproof

The following lemma is a standard version of Bernstein's inequality. 

\begin{lemma}[Bernstein's Inequality]\label{lem3}
If 
\begin{align*}
  \hat N = \sum_{t=0}^T I_t. 
\end{align*}
where $I_t$ are independent Bernoulli random variables and $n := \mathbb E [\hat N]$ then
\begin{align*}
  \mathbb P \left(
	\hat N - n \leq -x 
\right)
\leq 
\exp 
\left\{
	- \frac{x^2}{x+2n}
\right\}
\end{align*}
and, taking $x= n / 2$ gives,
\begin{align*}
  \mathbb P \Big( \hat N \leq \frac{n}{2} \Big) \leq e^{- \frac{n}{10}} 
\end{align*}
\end{lemma}

The following is a version of Chernoff's bound (or Hoeffding's bound).

\begin{lemma}[Chernoff-Hoeffding bound]\label{lem4}
If $\tilde r_a(n)$ is the empirical reward of arm $a$ after $n$ pulls and, recall, $r_a$ is the mean of arm $a$
then 
\begin{align*}
  \mathbb P \left(
	| \tilde r_a(n) - r_a | \geq \frac{\Delta}{3}
\right)
\leq 
2e^{- \frac{\Delta^2 n }{9}}
\end{align*}
\end{lemma}

\begin{proposition}\label{prop1}
	If we let $\hat r_a(T)$ be the empirical mean of arm $a$ at time $T$ and $r_a$ be the mean reward of arm $a$, then
\begin{align*}
  \mathbb P \left( \left|
	\hat r_a(T) - r_a
\right|
\geq \frac{\Delta}{3}
\right)
\leq 
\frac{C}{T^{\frac{1}{20 \alpha}}} 
+ 
\frac{D}{T^{\frac{\Delta^2}{36 \alpha}}}
\end{align*}
where $C$ and $D$ are constants depending on $\alpha, \Delta$ and $p_a(0)$ only.
\end{proposition}
\Beginproof
 As before, we let $\tilde r_a(n)$ be the empirical reward of arm $a$ after $n$ pulls.
We let $N_a(T)$ be the number of times arm $a$ is pulled by time $T$.
Notice that by Lemma \ref{lem1},
we know that the probability of pulling each arm is bounded below by $1/(2\alpha t + p_a(0)^{-1}) $. Thus $N_a(T)$ is stochastically bounded below by $\hat N(T)$ where 
\begin{align*}
  \hat N(T) = \sum_{t=0}^T I_t. 
\end{align*}
and $I_t$ are independent Bernoulli random variables with mean $1/(2\alpha t + p_a(0)^{-1}) $. We will apply Lemma \ref{lem3}, to this end we bound the mean of $\hat N(T)$ as follows
\begin{align*}
  n_T = \mathbb E [ \hat N(T)] 
&=
\sum_{t=0}^T 
	\frac{1}{2 \alpha t + p(0)^{-1}}
\\
&
\geq \int_0^{T-1} \frac{1}{2\alpha t + p(0)^{-1} } dt 
\\
&
\geq \int_1^T \frac{1}{2\alpha t + p(0)^{-1} } dt 
\\
&
=
\frac{1}{2 \alpha} \log \left(
	\frac{2 \alpha T + p(0)^{-1} }{2 \alpha + p(0)^{-1}}
\right)
\\
&
\geq 
\frac{1}{2\alpha} \log \left(
	\frac{2 \alpha T}{ 2\alpha + p(0)^{-1}}
\right) 
=
\frac{1}{2\alpha} \log T
+
\frac{1}{2\alpha} \log \left(
	\frac{2 \alpha }{ 2\alpha + p(0)^{-1}}
\right)  \, .
\end{align*}

Now we can bound our quantity of interest 
\begin{align*}
   & \mathbb P \left( \left|
	\hat r_a(T) - r_a
\right|
\geq \frac{\Delta}{3}
\right)
\\
=
&
\sum_{n=0}^T \mathbb P \Big(
	 N_a(T) = n ,
	\left|
	\tilde r_a(n) - r_a
\right|
\geq 
\frac{\Delta}{3}
\Big)
\\
= 
&
\sum_{n=0}^{\lfloor n_T/2\rfloor} \mathbb P \Big(
	 N_a(T) = n ,
	\left|
	\tilde r_a(n) - r_a
\right|
\geq 
\frac{\Delta}{3}
\Big)
+
\sum_{n=\lfloor n_T/2\rfloor +1}^T \mathbb P \Big(
	 N_a(T) = n ,
	\left|
	\tilde r_a(n) - r_a
\right|
\geq 
\frac{\Delta}{3}
\Big)
\\
\leq 
&
\mathbb P \Big(
	 N_a(T) \leq \frac{n_T}{2} 
\Big) 
+
\sum_{n=\lfloor n_T/2\rfloor +1}^T \mathbb P \left( \left|
	\tilde r_a(n) - r_a
\right|
\geq 
\frac{\Delta}{3}\right)
\\
\leq 
&
\mathbb P \Big(
	 \hat N(T) \leq \frac{n_T}{2} 
\Big) 
+
\sum_{n=\lfloor n_T/2\rfloor +1}^\infty 
2e^{- \frac{\Delta^2 n }{9}}
\\
\leq 
&
e^{- \frac{n_T}{10} } 
+
 \left[
	\frac{2e^{- \frac{\Delta^2}{9} } }{1-e^{- \frac{\Delta^2}{9}  }}
\right]
e^{- \frac{\Delta^2}{9} \frac{n_T}{2} }
\\
\leq 
&
\left[
\frac{2 \alpha }{ 2\alpha + p(0)^{-1}}
\right]^{-\frac{1}{20\alpha}}
\frac{1}{T^{\frac{1}{20 \alpha}}} 
+ 
\left[
\frac{2 \alpha }{ 2\alpha + p(0)^{-1}}
\right]^{-\frac{\Delta^2}{36\alpha}}
 \left[
	\frac{2e^{- \frac{\Delta^2}{9} } }{1-e^{- \frac{\Delta^2}{9}  }}
\right]
\frac{1}{T^{\frac{\Delta^2}{36 \alpha}}} \, .
\end{align*}
In the first inequality above, we use the fact that 
$\{N_a(T) = n, | \tilde r_a(n) - r_a | \geq \Delta /3  \} \subset \{N_a(T) = n  \}  $ and, similarly, $\{N_a(T) = n, | \tilde r_a(n) - r_a | \geq \Delta /3  \} \subset \{| \tilde r_a(n) - r_a | \geq \Delta /3 \} $.
In the 2nd inequality, we use the fact that $N_a(T)$ is stochastically bounded below by $\hat N(T)$. 
In the 3rd inequality, we apply the Chernoff bound, Lemma \ref{lem4}, and then let $T= \infty$ in 2nd the summation. 
In the 4th inequality, we apply Bernstein's inequality, Lemma \ref{lem3}, and sum the geometric series in the summation. 

From this we see the result above holds with
\begin{align*}
  C = \left[
\frac{2 \alpha }{ 2\alpha + p(0)^{-1}}
\right]^{-\frac{1}{20\alpha}}
\qquad
\text{and}
\qquad 
D = \left[
\frac{2 \alpha }{ 2\alpha + p(0)^{-1}}
\right]^{-\frac{\Delta^2}{36\alpha}}
 \left[
	\frac{2e^{- \frac{\Delta^2}{9} } }{1-e^{- \frac{\Delta^2}{9}  }}
\right]\, .
\end{align*}
\Endproof

\begin{proposition}\label{prop2}
	Let 
\begin{align*}
  T^\star = \max \left\{ T : |\hat r_a(T) - r_a | \geq \frac{\Delta}{3}, \text{ for some arm } a \right\}
\end{align*}
then, there exists a value $\alpha^\star$ such that for all  learning rates $\alpha < \alpha^\star$, it holds that
\begin{align*}
  \mathbb E [T^\star ] < \infty \, .
\end{align*}
\end{proposition}
\begin{remark}
$\bullet$	Notice that for all $t > T^\star$ it holds that $
  | \hat r_T(a) - \hat r_T(a^\star)| \geq \frac{\Delta}{3} 
$, $\forall a \neq a^\star$. Thus for $t > T^\star$ the probability of playing arms obeys the recursion 
\begin{align*}
  p_a(t+1) \leq p_a(t) - \frac{\Delta}{3} p_a(t)^2 \, \qquad \forall a \neq a^\star \, .
\end{align*}
$\bullet$ We note that from the proof that we take $\alpha^\star = \frac{\Delta^2}{72}$. The reader can note that this choice originates from the Chernoff bound in Lemma \ref{lem4} thus depends on the distance (relative entropy) between the sub-optimal arms and the optimal arm.

\end{remark}

\begin{proof}[Proof of Theorem \ref{ThrmAv}.]
We will make use of the bound that for $\eta>1$
\begin{align*}
  \sum_{s=t}^\infty \frac{1}{s^\eta} \leq  \frac{1}{t^{\eta}} +  \int_t^\infty \frac{1}{s^\eta} ds =  \frac{1}{t^{\eta}} + \frac{1}{(\eta-1)t^{\eta -1}}
\end{align*}

We assume that $\alpha$ is chosen so that $\frac{\Delta^2}{36 \alpha} > 2$ and $\frac{1}{20 \alpha } > 2$ 
  \begin{align*}
  \mathbb P \left( T^\star \geq t \right)
=
&
\mathbb P \left( \exists T \geq t \text{ and } a \text{ s.t. }  |\hat r_a(T) - r_a | \geq \frac{\Delta}{3} \right)
\\
\leq 
&
\sum_{a \in \mathcal A} 
 \sum_{T = t}^\infty 
 \mathbb P \left(
	\| \hat r_a(T) - r_a \| \geq \frac{\Delta}{3}
\right)
\\
\leq 
&
N 
\sum_{T=t}^\infty 
\left(
	\frac{C}{T^{\frac{1}{20 \alpha}}} 
+ 
\frac{D}{T^{\frac{\Delta^2}{36 \alpha}}}
\right)
\\
\leq 
&
N C \left(
	 \frac{1}{t^{\frac{1}{20\alpha}}} + \frac{1}{(\frac{1}{20\alpha}-1)t^{\frac{1}{20\alpha} -1}}
\right)
+
N D
\left(
	 \frac{1}{t^{\frac{\Delta^2}{36 \alpha}}} + \frac{1}{(\frac{\Delta^2}{36 \alpha}-1)t^{\frac{\Delta^2}{36 \alpha} -1}}
\right)\, .
\end{align*}
Now we can bound the expectation of $T^\star$:
\begin{align*}
  \mathbb E \left[
	T^\star
\right]
=
&
\sum_{t=0}^\infty \mathbb P(T^\star \geq t)
\\
\leq 
&
1+
\sum_{t=1}^\infty
\left[
	N C \left(
	 \frac{1}{t^{\frac{1}{20\alpha}}} + \frac{1}{(\frac{1}{20\alpha}-1)t^{\frac{1}{20\alpha} -1}}
\right)
+
N D
\left(
	 \frac{1}{t^{\frac{\Delta^2}{36 \alpha}}} + \frac{1}{(\frac{\Delta^2}{36 \alpha}-1)t^{\frac{\Delta^2}{36 \alpha} -1}}
\right)
\right]
\\
< 
&
\infty \, .
\end{align*}
The sum above is finite because each term is of the form $1/t^{\eta}$ for $\eta >1$ and thus has a finite sum.
\end{proof}

\Beginproof
The regret of the SAMBA algorithm with average rewards can be bounded with the following sequence of inequalities:
  \begin{align*}
  \mathcal R \! g(T) 
& =
\mathbb E \left[
	 \sum_{t=1}^T 
\sum_a \Delta_a p_t(a) 
\right]
\\
& \leq
\mathbb E \left[
	\sum_{t=1}^{T+T^\star} 
 \sum_a \Delta_a p_t(a) 
\right]
\\
&
=
\mathbb E \left[
	\sum_{t=1}^{T^{\star}}
\sum_a \Delta_a p_t(a) 
\right]
+
\mathbb E \left[
	\sum_{t=T^\star+1}^{T+T^\star} 
\sum_a 
	 \Delta_a p_t(a) 
\right]\\
&
\leq
N \mathbb E [ T^\star ] 
+
\sum_{t=1}^T \sum_a \Delta_a \frac{1}{1+\alpha \Delta t / 3  }
\\
&
\leq 
N\mathbb E \left[
	T^\star 
\right]
+
\sum_a \Delta_a \int_0^T \frac{1}{1+ \alpha \Delta t / 3} dt
\\
&
\leq 
N \mathbb E \left[
	T^\star 
\right]
+
 \frac{3 N }{\alpha \Delta} \log \left( 1+ \alpha \Delta T / 3 \right) 
\end{align*}
In the 1st inequality, we note that $T^\star$ is positive.
In the 2nd inequality, we note that $\Delta_a p_a(t) \leq 1$ for $t \leq T^\star$ and we note that $\tilde p(t) = p_{t+T^\star}$ obeys the recursion $\tilde p(t+1) \leq \tilde p(t) - \alpha \Delta \tilde p (t)^2 / 3$ for all $t$ and thus we apply Lemma \ref{lem2}. The remaining inequalities are standard bounds. 
\Endproof

%% file: SIMULATIONS.tex
\section{Simulation Study.}\label{Simulations}
The main contribution of the paper is a new proof that applies a blend of Stochastic Approximation and Markov chain techniques to bandit problems. The proof and algorithm likely extends to a broad class of sequential decision making problems.\footnote{Notions such as prior conjugacy, confidence intervals or even an arms cumulative reward may be intractable for general sequential decision making problems, such as reinforcement learning. Thus, it is useful to have stochastic approximation routine that can be applied to current rewards and has provable regret bounds for bandit problems.}
Nonetheless it is interesting to see how SAMBA behaves for bandit problems. This brief study is indicative of performance. We empirically compare SAMBA with and without cooling against a range of bandit algorithms: Thompson Sampling, UCB, Exp3, Gradient Bandit Algorithm (GBA), $\epsilon$-Greedy.

In each case there are hyper-parameter optimizations, tweaks and variants of the original SAMBA algorithm\footnote{E.g. Replacing randomized exploration with Metropolis-Hastings, ordering arms to do pairwise comparison to increase/decrease probabilities, applying different state dependent learning rates and applying different function approximations as discussed in the extensions section.} which help improve performance in different settings but perhaps the same could be said for UCB, Thompson Sampling and Exp3. We try to refrain from introducing numerous extensions and, instead, 
only consider preselected parameter choices and generic designs for each bandit algorithm. Code for this section is available on Github.\footnote{\url{https://github.com/neilwalton/MOR_Paper/blob/master/MOR_Bandits.ipynb}}

\subsection{SAMBA Learning Rate.}

We first analyze the dependence of SAMBA, with and without cooling, on its learning rate. This confirms the behavior proven in Theorem \ref{main_theorem} and Theorem \ref{3rd_theorem}. Also it helps us find reasonable parameter choices for both versions of SAMBA. 

\begin{figure}[!tbp]
	\centering
	\begin{minipage}{.5\textwidth}
		\centering
		\includegraphics[width=1.\linewidth]{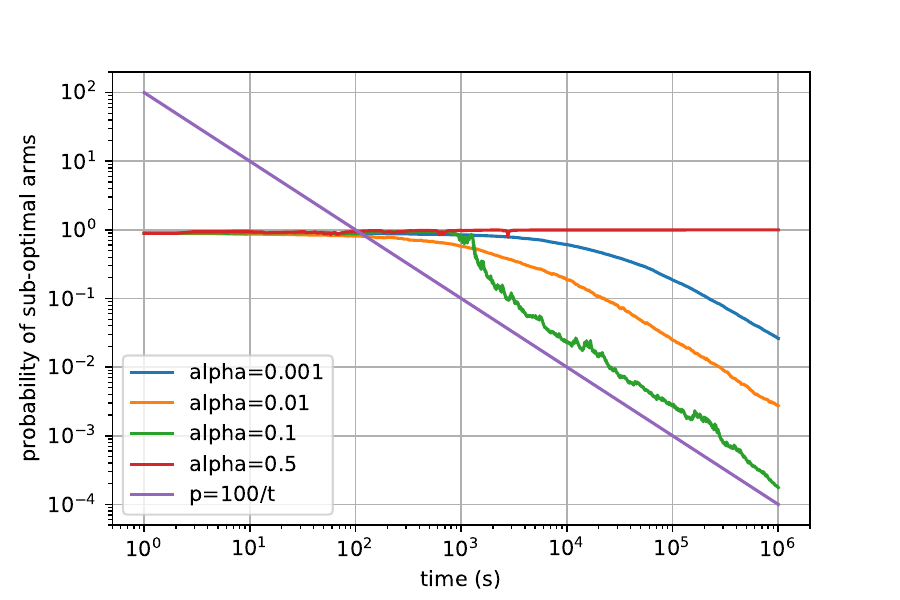}
		\caption{SAMBA without cooling.}
		\label{fig:Samba1}
	\end{minipage}%
	\begin{minipage}{.5\textwidth}
		\centering
		\includegraphics[width=1.\linewidth]{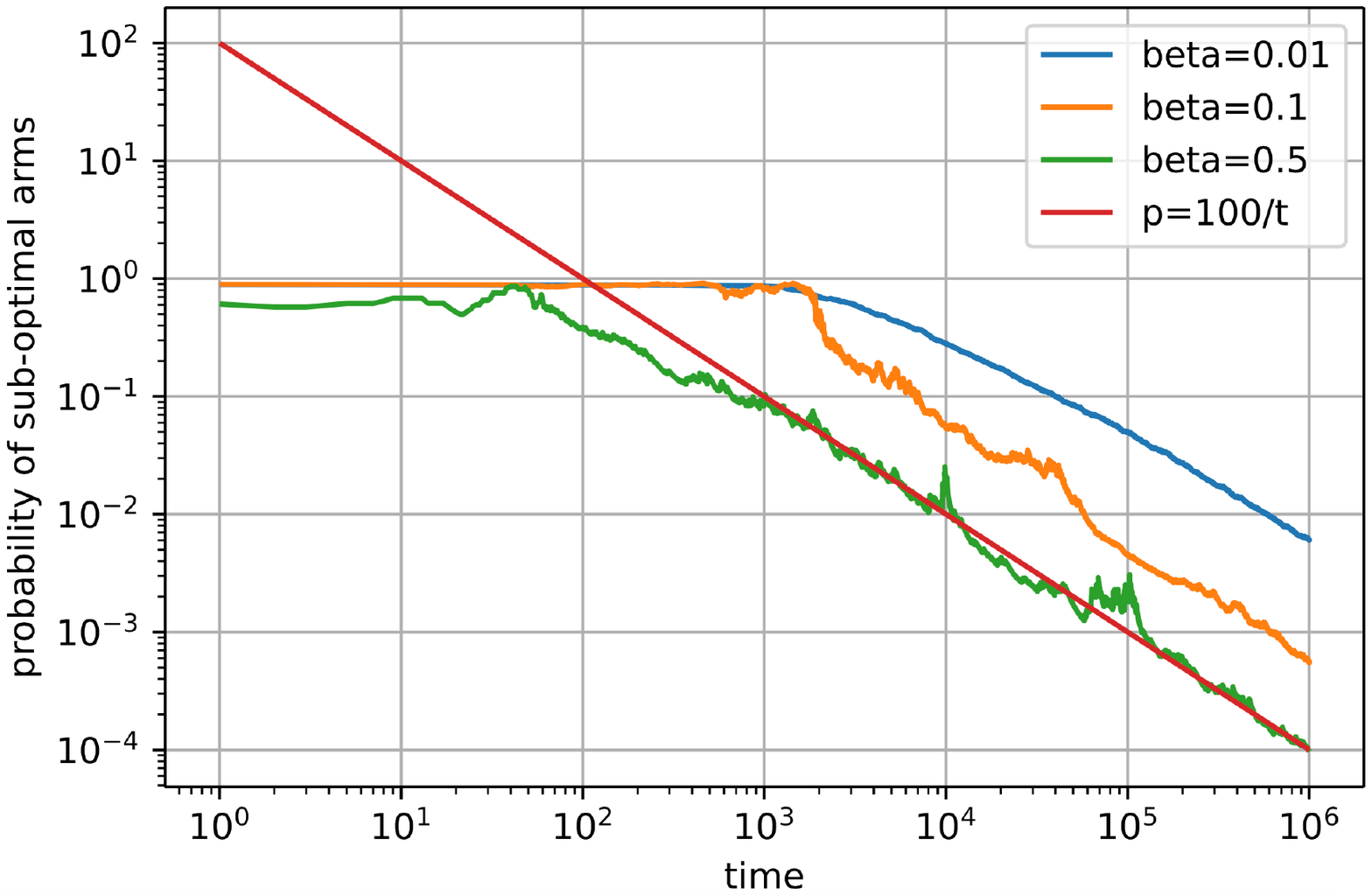}
		\caption{SAMBA with Cooling}
		\label{fig:Samba2}
	\end{minipage}
\end{figure}

In Figures \ref{fig:Samba1} and \ref{fig:Samba2} we apply SAMBA without cooling (as considered in Theorem \ref{main_theorem}) and with cooling (as considered in Theorem \ref{3rd_theorem}). Here the bandit problem has nine independent Bernoulli distributed arms each taking a probability of reward $r_a= 0.1, 0.2, ...,0.9$. We plot the probability of playing a sub-optimal arm on a log-log scale. We also plot $p = 100/t$ for reference, since $p\propto 1/t$ is required for logarithmic regret.

\begin{figure}[!tbp]
\centering
\includegraphics[width=0.5\linewidth]{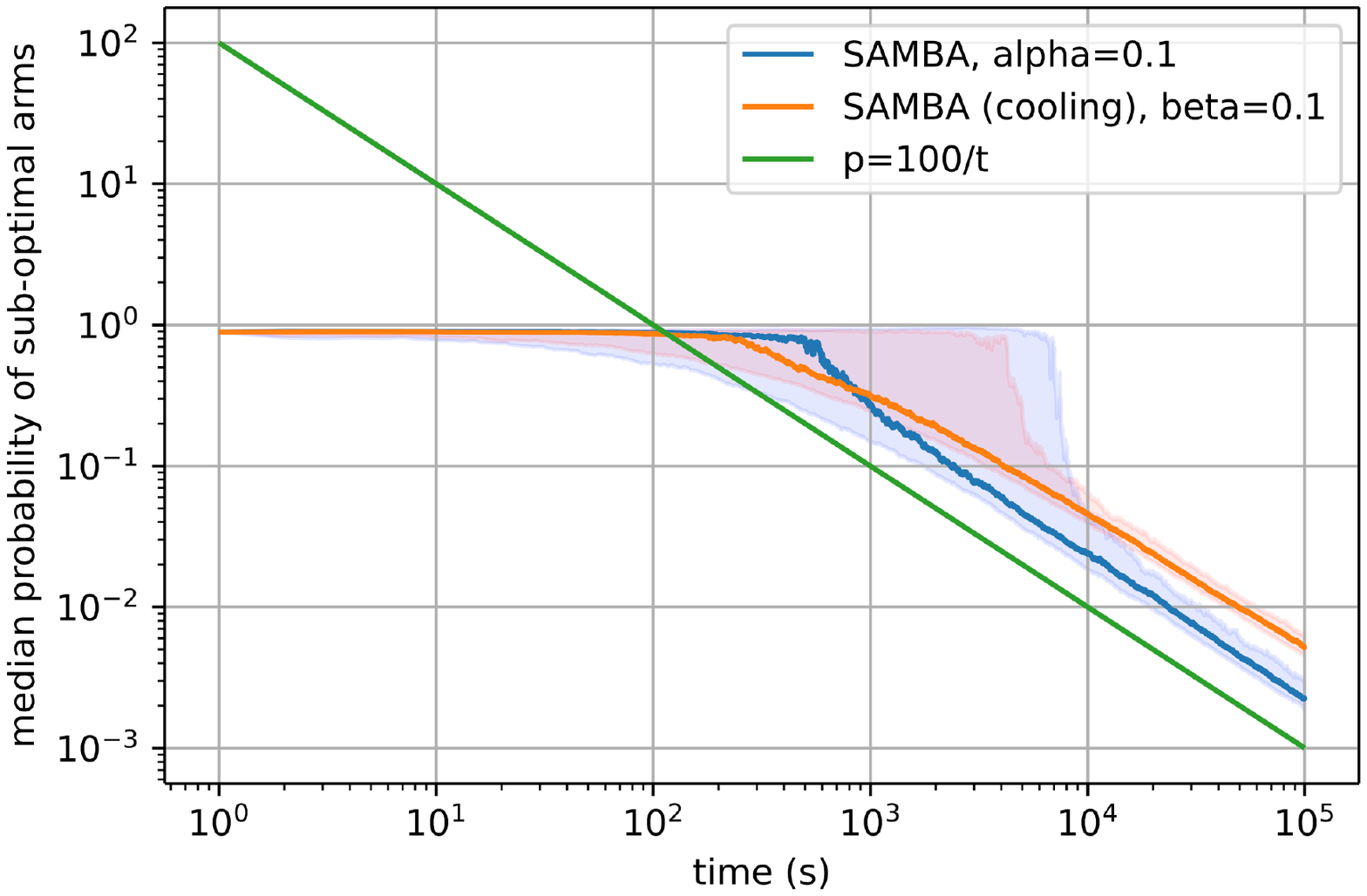}
\caption[caption]{{\tabular[t]{@{}l@{}}Median of SAMBA with and without cooling.  \\ (Shaded area is 10th to 90th percentile)\endtabular}}
	\label{fig:Samba3A}
\end{figure}

In Figure \ref{fig:Samba1}, we see for SAMBA without cooling that $\alpha$ behaves increasingly well upto $\alpha=0.1$. 
The slope of each line for $\alpha =0.1,0.01$ and $0.001$ is as expected for an algorithm with logarithmic regret. 
However, for $\alpha=0.5$ which is chosen too big so that condition \eqref{alpha_cond} is violated, the algorithm convergences on the arm with average reward $0.8$. Usually for $\alpha=0.5$ the algorithm selects the optimal arm, but this simulation instance demonstrates that a condition like \eqref{alpha_cond}  is required and also demonstrates the role of $\alpha$ as an error tolerance of the bandit algorithm. 
In Figure \ref{fig:Samba2}, we see for SAMBA with cooling that large values of $\beta$ remain convergent. 

Essentially, this first set of simulations reconfirms what we have already proved mathematically. Further, we find $\alpha=0.1$ for SAMBA and $\beta =0.1$ for SAMBA with cooling to be reasonably good parameter choices. In particular, this is indicated in Figure \ref{fig:Samba3A} where we plot the median performance for this parameter choice as well as the interval between the 10th and 90th percentile. We fix these parameter choices for the remainder of this simulation study.

\subsection{Comparison with Other Bandit Algorithms.}

We consider a four armed bandit problem with a moderate reward probabilities $r_a=0.1,0.5,0.8,0.9$ and small reward probabilities $r_a=0.01,0.05,0.08,0.09$. Here there are two arms with similar probabilities of reward and two arms that should be quickly established to be sub-optimal. We analyze regret over 1000 time steps and take the average over 1000 simulation runs.

In addition to SAMBA with $\alpha=0.1$ and SAMBA with cooling and $\beta = 0.1$, we consider Thompson Sampling with a uniform prior \citep{thompson1933likelihood}, UCB-1 from \citep{auer2002finite}, the Gradient Bandit Algorithm with $\alpha=0.1$ from \citep{sutton2018reinforcement}, Exp3 with $\eta=\sqrt{\log (N)/tN}$ from \citep{bubeck2012regret}, $\epsilon$-Greedy with decaying exploration $\epsilon=\min\{1,100/t \}$ and $\epsilon$-Greedy with $\epsilon =0.1$ from \citep{sutton2018reinforcement}.

\begin{figure}[!tbp]
	\centering
	\begin{minipage}{.5\textwidth}
		\centering
		\includegraphics[width=1.\linewidth]{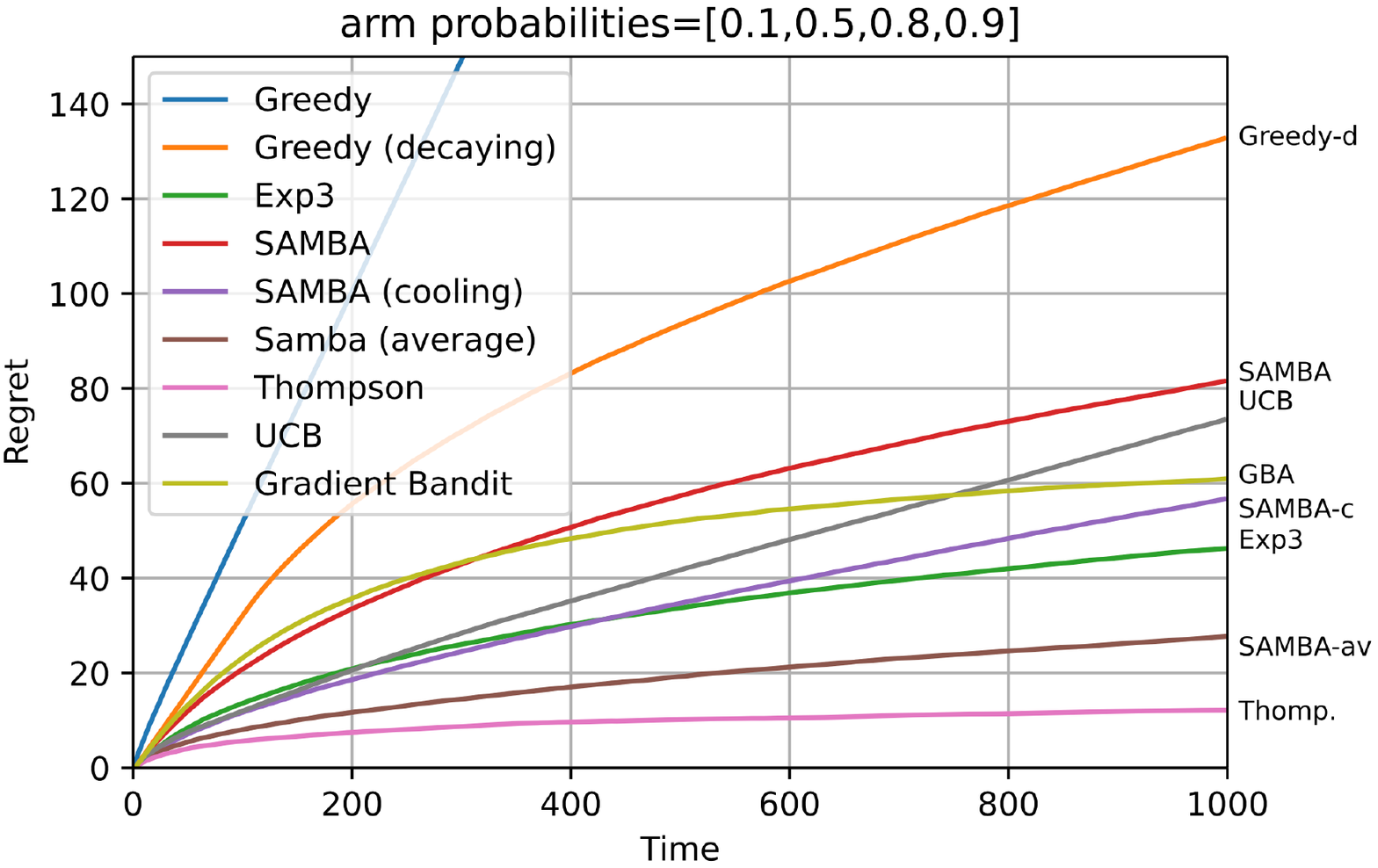}
		\caption{Bandit Comparison.}
		\label{fig:Samba3}
	\end{minipage}%
	\begin{minipage}{.5\textwidth}
		\centering
		\includegraphics[width=0.97\linewidth]{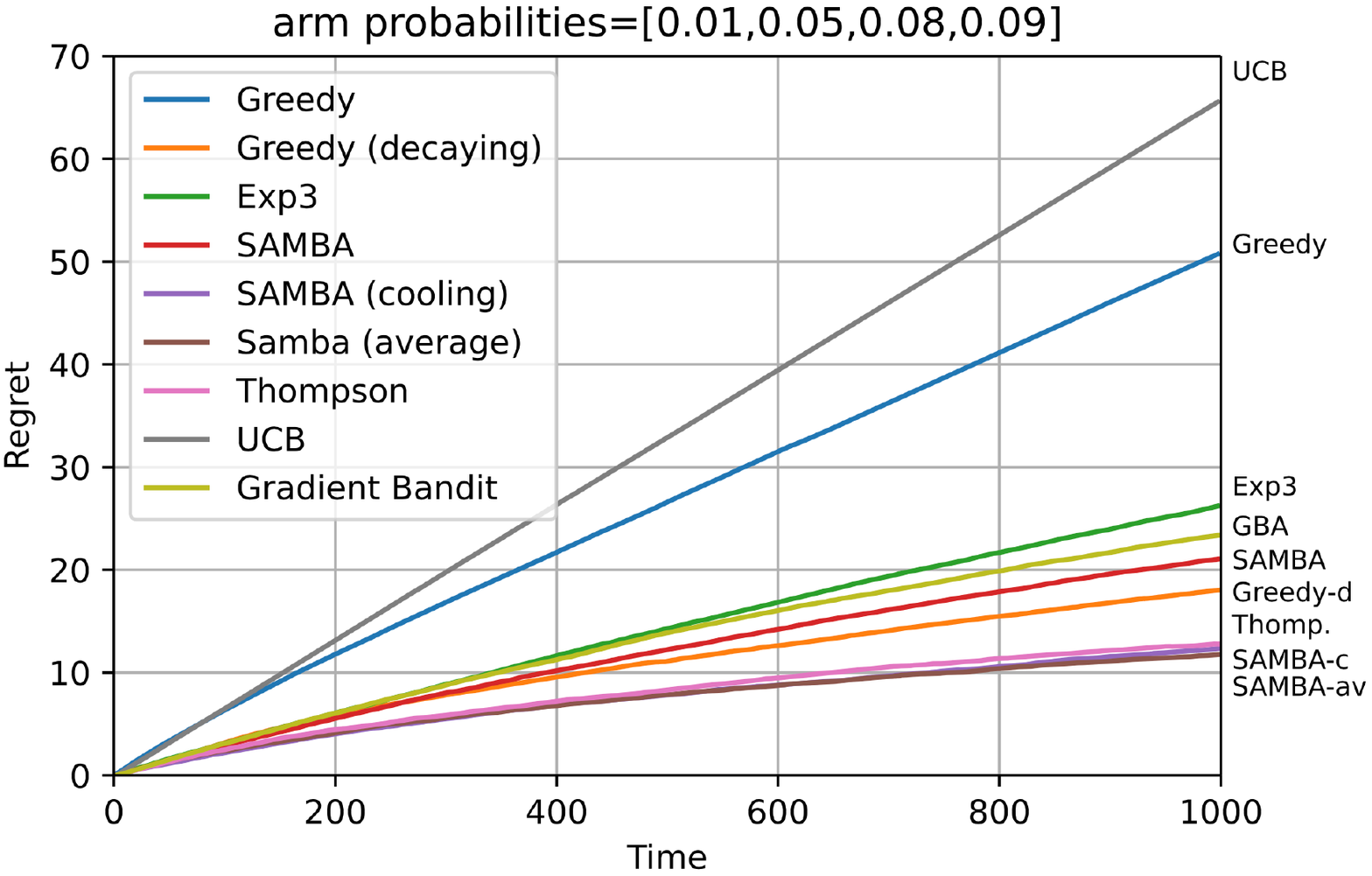} 
		\caption{Small Probabilities Comparison}
		\label{fig:Samba4}
	\end{minipage}
\end{figure}

In Figure \ref{fig:Samba3} and \ref{fig:Samba4}, we see that Thompson Sampling is the most effective method, though SAMBA with averaging has comparable performance. 
SAMBA with cooling out performs SAMBA. UCB performs worse for small reward probabilities, while both SAMBA algorithms improve relative to other methods. 

\subsection{Comparison with Large Numbers of Arms.}

Although UCB appears to perform worse in the previous experiments, its reward significantly improves when the number of arms increases. We consider a setting with $N=10,20,...,100$ arms. Here we consider $100$ experiments where the reward probability of each arm is drawn from a uniform distribution on $[0,0.1]$. Once this set of bandit problems is generated (and fixed), we apply each algorithm over $100,000$ time steps to each of the bandit problems and average the reward recieved.  This set up loosely mimics a sponsored search setting where ad click-through rates are roughly in the range $[0.001, 0.1]$ and tens of ads within a given category (or query) vie for thousands of impressions.
\begin{figure}
	\centering
	\includegraphics[width=0.6\linewidth]{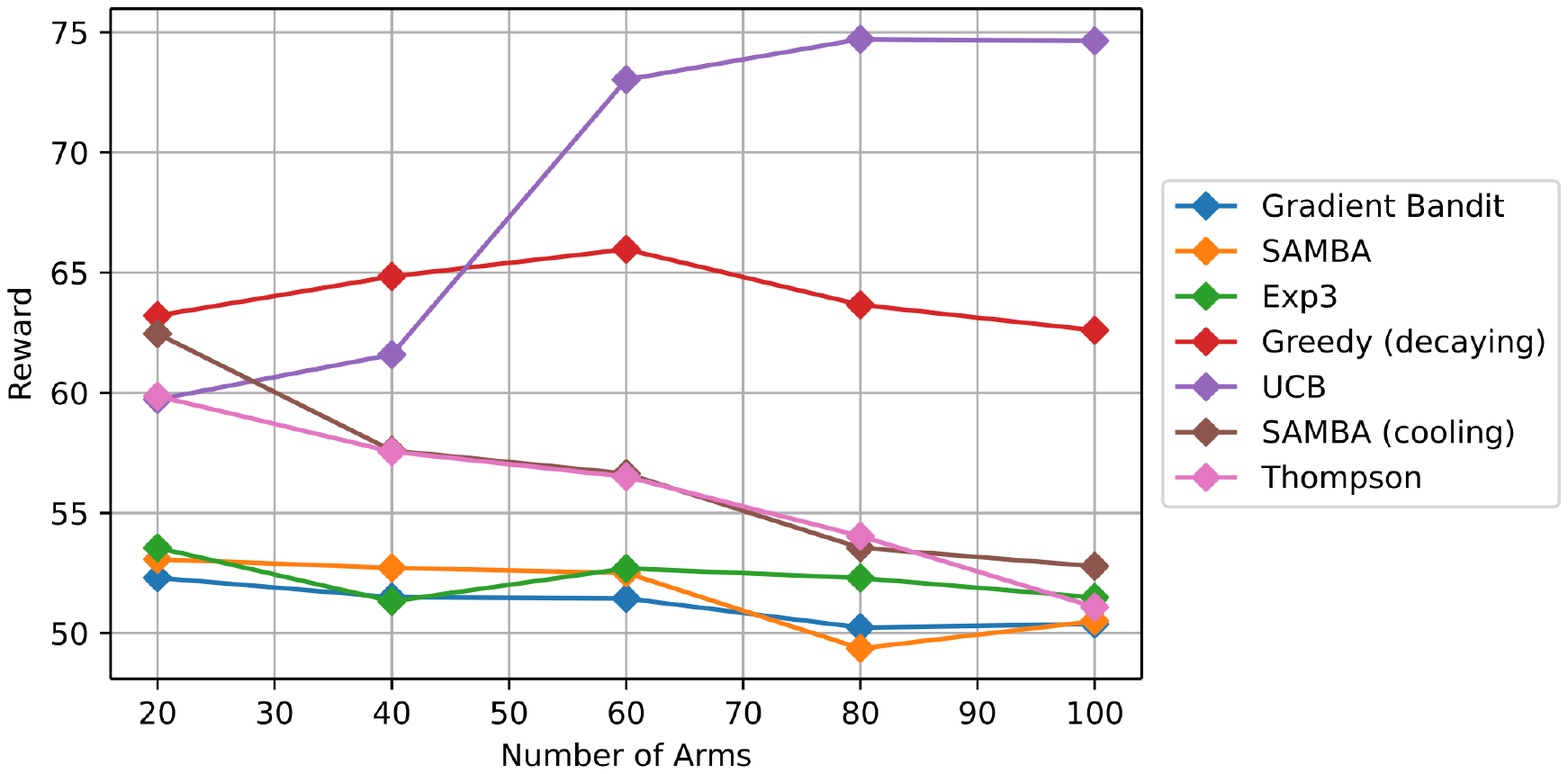}
	\caption{Dependence on number of arms.}
	\label{fig:Samba5}
\end{figure}

In Figure \ref{fig:Samba5}, we find that UCB and $\epsilon$-Greedy with decaying $\epsilon$ perform well. 
The performance of Thompson Sampling degrades, as do other policies. 
Here UCB has the advantage that it ensures exploration according to a non-random rule, and thus is more systematic in checking each arm. 
SAMBA with cooling performs the best after UCB and $\epsilon$-Greedy. Further increasing its learning rate $\beta$ can further help it quickly eliminate underperforming arms. 

\subsection{Summary of Simulations.}

SAMBA performs better with cooling than without, because it is more aggressive at initial exploration. 
Broadly we find Thompson Sampling is the best performing algorithm (which advocates the advantages of prior conjugacy) except when there is a large number of arms where UCB performs well (which advocates the advantages of non-random exploration). 
SAMBA and GBA have comparable performance.
This is, perhaps, not surprising since they belong to same class of algorithms. 
SAMBA with averaging is very competitive with all the bandit algorithms that we have studied. 
Broadly, we find SAMBA algorithms to have good performance that is in-line with more established multi-arm bandit algorithms. 

%% file: CONCLUSION.tex
\section{Conclusion.}\label{sec:Conc}

We show a combination of Markov chain and martingale analysis can be applied to prove convergence for a policy gradient algorithm applied to a multiarm bandit problem. As the rates of convergence found in deterministic models may not lead to sufficient exploration when applied to stochastic policy gradient algorithms, we emphasis the importance of appropriate step sizes to ensure convergence and low regret. 


A natural extension to apply the approach to reinforcement learning. For example a tabular reinforcement learning update would give an update of the form:
\begin{align*}
    p({a | s}) \leftarrow p({a|s}) + \alpha p({a|s})^\delta 
\Big( \bar{Q}_{s,a}-\max_{a'} \bar{Q}_{s,a'} \Big).
\end{align*}
where $\bar{Q}_{s,a}$ is an estimate of the $Q$-factor of state $s$ and action $a$, $p(a|s)$ is the probability of choosing arm $a$ in state $s$, and $\delta$ is a constant. The theoretical analysis of this procedures  remains open.

%% file: APPENDIX.tex
\section{Appendix}


\subsection{Proof of Corollary \ref{prop1}.}
\label{corol5}
Lemma \ref{half_lemma} shows that if $p_{a^\star}$ is not too small then there is always a positive probability of reaching a state with $p_{a^\star} >\frac{1}{2}$.

\begin{lemma}\label{half_lemma}
	If $p_{a^\star}(0) \geq \frac{1}{\hat x}$ then there exists $n\in\mathbb N$ such that 
	\[
	\mathbb P
	\Big(
	p_{a^\star}(n)
	>
	\frac{1}{2}
	\, ,\,
	p_{a^\star}(t) \geq \frac{1}{\hat x}\;\; \forall t \leq n
	\Big)
	\geq 
	\left(
	\frac{r^\star}{\hat x}
	\right)^n\, .
	\]
\end{lemma}
Lemma \ref{half_lemma} analyses the probability of a run of rewards of $1$ on arm $a^\star$. We give a proof for when the learning rate has the form $\gamma (p_a) = \alpha p_a^2$ (as required in Theorem \ref{main_theorem}) in Section \ref{proof:half_lemma}  and where $\gamma(p_a) = \beta p_a^2/\log(e-\log(p_a))$ (as required in Theorem \ref{3rd_theorem}) in Section \ref{proof_2:half_lemma}. 

Lemma \ref{technical_lemma} is a technical lemma. It shows that a finite expected recurrence time to some set $B$ with a positive probability of reaching some set $C$ implies $C$ has a finite expected  recurrence time. 

\begin{lemma}\label{technical_lemma}
	If $( X(t) : t\in\mathbb Z_+)$ is a Markov chain on $\mathcal X \subset \mathbb R^d$ and $A$, $B$ and $C$ are disjoint subsets of $\mathcal X$ such that 
	for some $n\in \mathbb N$ and $\delta >0$ and $K < \infty$
	\begin{subequations}
		\begin{align}
		\inf_{x_0 \in B}
		\mathbb P 
		( X(n) \in C, X(k) \notin A,\,  \forall  k < n  | X(0) = x_0 )  
		& \geq \delta 
		\label{rec:state}
		, 	\\
		\sup_{x_0 \in B, x_1 \in A}
		\mathbb E 
		[\tau_B 
		|
		X(1) = x_1,
		X(0) = x_0 
		]
		&
		< K
		\label{rec:time}
		\end{align}	
	\end{subequations}
	where $\tau_B = \min\{t \in \mathbb Z_+ : X(t) \in B \}$ 
	then $\tau_C := \min\{t \in \mathbb Z_+ : X(t) \in C \}$ is such that
	\[
	\sup_{x_0 \in B}
	\mathbb E 
	[\tau_C 
	|
	X(0) = x_0 
	]
	< \infty \, .
	\]
\end{lemma}
The argument for the above lemma is essentially as follows, whenever the process $X$ is in $B$ then there is a positive probability of $\delta$ that we are in set $C$ in $n$ units of time. Thus there are at most $n$ plus a geometrically distributed (parameter $\delta$) number of times that the Markov chain can be in $B$ before visiting $C$. Since the time between each step of the Markov chain is $B$ has expectation bounded above by $K$ then the expected time is less than $ K (n + 1 / \delta )$. 
Although this description is probably sufficient, we give the more formal argument in the appendix.

 \Beginproof
	The proposition follows by applying Lemma \ref{technical_lemma}. Specifically we let 
	\begin{align*}
	A
	=
	\Big\{ \bm p \in \mathcal P : p_{a^\star} < \frac{1}{\hat x} \Big\}
	\, ,
	\quad
	B
	=
	\Big\{ 
	\bm p\in \mathcal P : 
	\frac{1}{\hat x} 
	\leq 
	p_{a^\star} 
	\leq  
	\frac{1}{2}
	\Big\}
	\,
	,
	\quad
	C 
	=
	\Big\{ 
	\bm p\in \mathcal P : 
	p_{a^\star} 
	>  
	\frac{1}{2}
	\Big\}\, .
	\end{align*}
	By Proposition \ref{first_lemma} 
	\[
	\sup_{\bm p^{(0)} \in B, \bm p^{(1)} \in A}
	\mathbb E 
	[\tau_B 
	|
	\bm p(1) = \bm p^{(1)},
	\bm p(0) = \bm p^{(0)} 
	]
	\leq 
	1+
	\sup_{p\in E(\hat x)} [\tau(\hat x)
	|
	\bm p(0) = \bm p
	]
	< \infty\, .
	\]
	By Lemma \ref{half_lemma}
	\[
	\mathbb P
	\Big(
	\bm p(n) \in C,\,\,
	\bm p(t) \notin A \,\, 
	\forall t \leq n
	\Big)
	\geq 
	\left(
	\frac{r^\star}{\hat x}
	\right)^n .
	\]
	Thus conditions \eqref{rec:state} and 
	\eqref{rec:time} of Lemma \ref{technical_lemma} are satisfied. So by Lemma \ref{technical_lemma} we have that 
	\[
	\sup_{\bm p \in E(2) } \mathbb E [ \tau | \bm p(0) = \bm p] 
	\leq 
	\sup_{\bm p \in B } \mathbb E [ \tau | \bm p(0) = \bm p] < \infty \, .
	\]
	   
\Endproof

\subsection{Proof of Lemma \ref{half_lemma} with $\gamma(p_a)=\alpha p_a^2$.}\label{proof:half_lemma}

\begin{proof}[Proof of Lemma \ref{half_lemma}.]
	Given $p_{a^\star}(0) > \frac{1}{\hat x}$, the probability $a^\star$ is played and receives a reward of $1$ is bounded below by $r^\star/\hat x$. Since $p_{a^\star}$ increases each time that it receives a reward of $1$. The probability of arm $a^\star$ being played and receiving a reward of $1$ for each of the next $n$ steps is bounded below by $(r^\star/\hat x)^n$. 
	
	Also, since $p_{a^\star}$ is increasing along this sequence, there is a value $t^\star$ [which we will bound above shortly] such that $a^\star \neq a_{\star}(t)$ for $t < t^\star$ and $a^\star = a_{\star}(t)$ for $t \geq t^\star$.
	
	For $t=0,...,t^\star-1$, $a^\star \neq a_{\star}(t)$  so by update \eqref{update}
	\[
	p_{a^\star}(t) 
	=
	(1+\alpha) p_{a^\star}(t-1)
	=
	...
	=
	(1+\alpha)^t p_{a^\star}(0)
	\geq {(1+\alpha)^t  }/{\hat x}\, .
	\]
	Thus $a^\star = a_{\star}(t)$ must hold for all $t$ such that 
	$
	(1+\alpha)^t / \hat x > 1/2\, 
	$.
	Therefore,
	\begin{equation}
	\label{t_star}
	t^\star \leq 
	\bigg\lceil
	\frac{\log (\hat x/2 )}{\log (1+\alpha)}
	\bigg\rceil \, .	
	\end{equation}
	For $t \geq t^\star$ and $a\neq a^\star$,
	\[
	p_a(t+1)
	=
	p_a(t)
	-
	\alpha
	\frac{p_a(t)^2}{p_{a^\star}(t)}
	\leq 
	p_a(t) 
	- \alpha
	p_a(t)^2\, .
	\]
	By Lemma \ref{upperbound},
	\[
	p_a(t) 
	\leq 
	\frac{
		p_a(t^\star) 
	}{
		1+ \alpha p_a(t^\star) ( t- t^\star)
	}
	\leq 
	\frac{1}{N+\alpha (t-t^\star)}\, .
	\]
	The second inequality above holds since $p_a(t^\star) \leq \frac{1}{N}$. Summing over $a\neq a^\star$ gives
	\[
	q_{a^\star}(t) 
	\leq 
	\frac{N-1}{N+ \alpha (t-t^\star)}\, .
	\]
	Thus $q_{a^\star}(n) < \frac{1}{2}$ for $n$ such that 
	\[
	\frac{N-1}{N + \alpha( n-t^\star)}
	\leq \frac{1}{2}\, ,
	\]
	or equivalently 
	$
	t^\star + {(N-2)}/{\alpha}  \leq n\, .
	$
	Thus by \eqref{t_star}, $p_{a^\star}(n) > \frac{1}{2}$ for all $n$ such that 
	\[
	n=  \bigg\lceil
	\frac{\log (\hat x/2 )}{\log (1+\alpha)}
	\bigg\rceil 
	+
	\left\lceil \frac{N-2}{\alpha} \right\rceil 
	+1\, .
	\]
   
\Endproof

\subsection{Proof of Lemma \ref{half_lemma} with $\gamma(p_a)=\beta p_a^2/\log ( e - \log p)$.}
\label{proof_2:half_lemma}
As discussed the proof is very similar to Lemma \ref{half_lemma} in Section \ref{proof:half_lemma}. The two differences are that we need to bound $\alpha$ below as it is now changing, and, instead of applying Lemma \ref{upperbound}, we need to apply Lemma \ref{upperbound_2_III}.

%

\begin{proof}[Proof of Lemma \ref{half_lemma}.]
Given $p_{a^\star}(0) > \frac{1}{\hat x}$, the probability $a^\star$ is played and receives a reward of $1$ is bounded below by $r^\star/\hat x$. Since $p_{a^\star}$ increases each time that it receives a reward of $1$. The probability of arm $a^\star$ being played and receiving a reward of $1$ for each of the next $n$ steps is bounded below by $(r^\star/\hat x)^n$. 

Also, since $p_{a^\star}$ is increasing along this sequence, there is a value $t^\star$ [which we will bound above shortly] such that $a^\star \neq a_{\star}(t)$ for $t < t^\star$ and $a^\star = a_{\star}(t)$ for $t \geq t^\star$. Also, since $p_{a^\star}(t) \geq \hat x^{-1}$, we have that $\alpha(p_{a^\star}(t)) \geq \alpha( \hat x^{-1})$ for $\alpha(p) := \beta/\log ( e - \log p)$.

For $t=0,...,t^\star$, $a^\star \neq a_{\star}(t)$  so by update \eqref{update}
\[
p_{a^\star}(t) 
\geq
(1+\alpha(\hat x^{-1})) 
\cdot p_{a^\star}(t-1)
=
...
=
(1+\alpha(\hat x^{-1}))^t 
\cdot p_{a^\star}(0)
\geq {(1+\alpha(\hat x^{-1}))^t  }/{\hat x}\, .
\]
Thus $a^\star = a_{\star}(t)$ must hold for all $t$ such that 
$
(1+\alpha(\hat x^{-1}))^t / \hat x > 1/2\, 
$.
Therefore,
\begin{equation}
\label{t_star_2}
t^\star \leq 
\bigg\lceil
\frac{\log (\hat x/2 )}{\log (1+\alpha(\hat x^{-1}))}
\bigg\rceil =: t_0 \, .	
\end{equation}
For $t \geq t^\star$ and $a\neq a^\star$,
\begin{align*}
p_a(t+1)
=
p_a(t)
-
\alpha(p_a(t))
\frac{p_a(t)^2}{p_{a^\star}(t)}
&
\leq 
p_a(t) 
- \alpha (p_a(t))
p_a(t)^2\, 
=
p_a(t) 
- \frac{\beta p_a(t)^2}{ 1-\log(p_a(t))}\, .
\end{align*}
Applying the 2nd bound in  Lemma \ref{upperbound_2_III},
\[
p_a(t) 
\leq 
\frac{1}{\beta (t-t^\star)}
\log\left( e + \log (\beta (t-t^\star))\right)
\]
for $\beta (t-t^\star) \geq 1 $. Summing over $a\neq a^\star$ gives
\[
q_{a^\star}(t) 
\leq 
\frac{N}{\beta (t-t^\star)}
\log\left( e + \log (\beta (t-t^\star))\right)
.
\]
Thus $q_{a^\star}(n) < \frac{1}{2}$ for $n \geq t_0$ (with $t_0$ defined above) and such that $\beta (n-t_0) \geq 1 $ and
\[
\frac{N}{\beta (n-t_0)}
\log\left( e +  \log (\beta (n-t_0))\right)
\leq \frac{1}{2}\, .
\]
\hfill $\square$\endproof

\subsection{Proof of Lemma \ref{technical_lemma}.}

\begin{proof}[Proof of Lemma \ref{technical_lemma}.]
	We let $\hat X(s)$ be the process that $X(t)$ follows when it is inside the set $B$ (and $A$). Specifically we let 
	$\tau_A^{(0)} = \tau_B^{(0)} = \tau_B^{-1} = 0$ and 
	\begin{align*}
	\tau_{A}^{(k)} 
	&
	=
	\min 
	\left\{
	t > \tau^{(k-1)}_B 
	:
	X(t) 
	\in A
	\right\},
	\\
	\tau_B^{(k)}
	&
	=
	\min 
	\left\{
	t > \tau_A^{(k)}
	:
	X(t)
	\in B 
	\cup C
	\right\}.
	\end{align*}
	and we define
	\begin{align*}
	\hat X (s) 
	=
	X 
	(t_s)
	\quad
	&
	\text{ where }
	\quad 
	t_s 
	=
	s
	+
	\sum_{i=0}^k
	\tau_B^{(i)}
	-\tau_A^{(i)}
	\\
	&\text{ for }\qquad 
	s\in \left[ 
	\sum_{i=0}^k 
	\tau_A^{(i)}
	-\tau_B^{(i-1)}
	,
	\sum_{i=0}^{k+1} 
	\tau_A^{(i)}
	-\tau_B^{(i-1)}
	\right)\, .	
	\end{align*}
	Notice 
	\begin{equation*}
	t_{s+1} =
	\begin{cases}
	t_s +1 
	& 
	\text{ if } 
	X(t_s+1) \in B \cup C\, ,\\
	t_s + \tau_B^{(i)} - \tau_A^{(i)} \, ,
	&
	\text{ if }
	X(t_s+1) \in A,\text{ and, thus, }t_s = \tau^{(i)}_A\text{ for some }i \, .
	\end{cases}
	\end{equation*}
	Thus, as $\mathbb E [ t_{s+1} - t_s | X(t_s)] < K $, by the Markov property and \eqref{rec:time}.
	
	Further we define 
	\[
	\sigma_C
	:=
	\min \{ s : \hat X (s) \in C \}
	\]
	By \eqref{rec:state},
	\begin{equation*}
	\mathbb P
	(
	\hat X(s+n) \in A 
	|
	\hat X(s) \in B
	)
	> \delta 
	\end{equation*}
	Thus
	\[
	\mathbb E_{x_0} [ \sigma_C ] \leq \frac{1}{\delta} + n\, .
	\]
	So we can bound the number of steps the chain $\hat X(s)$ takes to reach $C$ . Since each unit step of $\hat X(s)$ is $\tau_{s+1} -\tau_s$ steps for $X(t)$, we have that
	\[
	\tau_C 
	=
	\sum_{s=0}^{\sigma_C-1}  t_{s+1} - t_s 
	\]
	Thus,
	\begin{align*}
	\mathbb E [\tau_C ]
	&
	=
	\mathbb E \bigg[
	\sum_{s=0}^{\sigma_C-1}
	t_{s+1} - t_s 
	\bigg]
	=
	\mathbb E \bigg[
	\sum_{s=0}^{\infty}
	(t_{s+1} - t_s )
	\mathbb I
	[\sigma_C > s]
	\bigg]
	\\
	&
	=
	\mathbb E \bigg[
	\sum_{s=0}^{\infty}
	\mathbb E \big[
	t_{s+1} - t_s 
	|
	X(t_s)
	\big]
	\mathbb I
	[\sigma_C > s]
	\bigg]
	\\
	&
	\leq 
	K
	\mathbb E
	\left[
	\sum_{s=0}^\infty 
	\mathbb I [ \sigma_C > s]
	\right]
	=
	K\mathbb E [ \sigma_C]
	\\
	&
	\leq K \left( \frac{1}{\delta} + n 
	\right)
	\end{align*}
	as required.
   
\Endproof

%

\subsection{Proof of Lemma \ref{Regret_Lemma}.}
\begin{lemma}
	\label{Regret_Lemma}
	\begin{equation*}
	\mathcal R\! g(T)
	=
	\sum_{a: a\neq a^\star}
	(r^\star - r_a) 
	\mathbb E 
	\left[
	\sum_{t=0}^{T-1} p_a(t) 
	\right]\, .
	\end{equation*}
\end{lemma}

\proof{Proof of Lemma \ref{Regret_Lemma}.}
	The following calculation uses standard properties of the conditional expectation and probabilities, see \citet{williams1991probability}. A similar calculation is given in \citet{lai1985asymptotically}.
	\begin{align*}
	\mathcal R\! g(T)&:= r^\star T - \mathbb E \left[ \sum_{t=0}^{T-1} \sum_{a\in \mathcal A} I_a(t) R_a(t) \right]
	\\
	& =
	\sum_{t=0}^{T-1}
	r^\star 
	- \mathbb E \left[ 
	\sum_{t=0}^{T-1} 
	\sum_{a\in \mathcal A} 
	\mathbb E [
	I_a(t) R_a(t) 
	| H({t-1}) ]
	\right]
	\\
	&
	=
	\mathbb E 
	\left[
	\sum_{t=0}^{T-1}
	r^\star 
	-
	\sum_{t=0}^{T-1}
	\sum_{a\in\mathcal A}
	r_a p_a(t)
	\right]
	\\
	&
	=
	\mathbb E 
	\left[
	\sum_{t=0}^{T-1}
	r^\star 
	(1-p_{a^{\star}}(t))
	-
	\sum_{t=0}^{T-1}
	\sum_{a\neq a^\star}
	r_a p_a(t)
	\right]
	\\
	&
	=
	\mathbb E 
	\left[
	\sum_{t=0}^{T-1}
	\sum_{a\neq a^\star}
	r^\star 
	p_a(t)
	-
	\sum_{t=0}^{T-1}
	\sum_{a\neq a^\star}
	r_a 
	p_a(t)
	\right]
	\\
	&
	=
	\sum_{a\neq a^\star}
	(
	r^\star 
	-r_a
	)
	\mathbb E 
	\left[
	\sum_{t=0}^{T-1}
	p_a(t)
	\right]\, .
	\end{align*}	
	The 2nd equality, uses the tower property of the conditional expectation. The 3rd equality applies the definition of $p_a(t)$ and applies the r\^{o}le of independence to $R_a(t)$. The 5th equality, uses that $1-p_{a^\star}(t) = \sum_{a\neq a^\star} p_a(t)$.
\Endproof

\subsection{Proof of Lemma \ref{LemABC}.}
The following is a technical lemma.

\begin{lemma}\label{LemABC} For positive constants $A,B$ and $C$ with
	$B > 2C$  and  $T\in\mathbb N$
	\[
	\sum_{s=0}^{T-1} 
	\frac{A}{B+Cs} 
	\leq 
	\frac{A}{C} 
	\log T 
	\, .
	\]
\end{lemma}

 \Beginproof
	The proof holds by the standard method of bounding a sum above  by its integral:
	\begin{align*}
	\sum_{s=0}^{T-1} 
	\frac{A}{B+Cs}
	&=
	\frac{A}{C}
	\sum_{s=0}^{T-1} 
	\frac{1}{B/C+s}
	\le 
	\frac{A}{C}
	\sum_{s=0}^{T-1} 
	\frac{1}{2+s}\\
	&\leq 
	\frac{A}{C}
	\int_{0}^{T-1} \!\!
	\frac{1}{1+ s}
	ds
	=\frac{A}{C}\log T.
	\end{align*}
\Endproof

\subsection{Proof of Lemma \ref{lemma3}.}
\label{app:lemma3}

\begin{lemma}\label{lemma3}
	If $( p_a(t) : a\in\mathcal A)$ is such that $p_a(t) >0$ and $\sum_{a} p_a(t) = 1$ then, under the update \eqref{update}, $( p_a(t+1) : a\in\mathcal A)$ is such that $p_a(t+1) >0$ and $\sum_a p_a(t+1) = 1$.
\end{lemma}

 \Beginproof
	It is immediate that $\sum_a p_a(t+1) = 1$, since the terms added to $p_a(t)$ in update \eqref{update}.
	
	We now show that $p_a(t+1)>0$ for all $a\in \mathcal A$. If $I_a(t) = 1$ for some $a\neq a_\star(t)$ then the only probability that decreases is $p_{a_\star} (t)$. Thus since $R_a(t)\leq 1$
	\[
	p_{a_\star(t)}(t+1)
	=
	p_{a_\star(t)}(t)
	-
	\alpha
	p_{a_\star(t)}(t)
	R_a(t)
	\geq
	p_{a_\star(t)}(1-\alpha)
	>
	0\, .
	\]
	So, in this case, all components of $\bm p(t+1)$ are positive.
	
	If $ I_\star (t):=I_{a_\star(t)}  = 1$ then $p_a(t)$ decreases for each $a\neq a_\star(t)$. In this case, since $R_\star(t) \leq 1$ and $p_a(t) \leq p_{a_\star(t)}(t)$, we have that
	\[
	p_a(t+1)
	=
	p_a(t)
	-
	\alpha
	p_a(t)^2
	\frac{R_\star(t)}{p_{a_\star(t)}(t)}
	\geq 
	p_a(t)
	-
	\alpha
	p_a(t)
	\frac{p_a(t)}{p_{a_\star(t)}(t)}
	\geq
	p_a(t) (1-\alpha)\, .
	\]
	So each case, we have that $\bm p(t+1)$ is positive, as required.	   
\Endproof

%% file: THEOREM_3.tex
\section{Proof of Theorem \ref{3rd_theorem}.}\label{sec:3rd_proof}

We do not require a gap dependent learning rate but at a small cost on our regret bound in Theorem \ref{main_theorem}.  This is stated in Theorem \ref{3rd_theorem}. The theorem demonstrates that we can gain a regret bound of order $O(\log T \log \log T)$. Consequently the regret bound is $O((\log T)^{1+\epsilon})$ for any $\epsilon>0$. Thus emphasizes the point that the gap to optimal rate $\log T$ can be made arbitrarily small. 

The proof follows a similar pattern to Theorem \ref{main_theorem}. Some results are identical. These results we state but refer the reader to the earlier proof. The main changes required are 
to replace Lemma \ref{upperbound} with Lemma \ref{upperbound_2_III}, Proposition \ref{first_lemma} with Proposition \ref{2nd_lemma_III}, Proposition \ref{LemExpectq} with Proposition \ref{LemExpectq_2_III}. 
 We state and prove these results. We also restate results whose proof is unchanged and refer the reader to the appropriate results in the main body of the paper.

Lemma \ref{upperbound_2_III} below is a discrete-time analysis of the o.d.e. $\dot{q}(t) = \eta q(t)^2/(\log(e-\log q(t)))$.

\begin{lemma}
	\label{upperbound_2_III}
	If $\bar q(t)$ is a sequence of real numbers belonging to the interval $(0,1)$ such that 
	\begin{equation}\label{qbounds_III}
	\bar q (t+1)
	-
	\bar q (t)
	\leq 
	-
	\frac{\eta \bar q(t)^2}{
	\log (e - \log ( \theta \bar q(t)) )
	}	
	\end{equation}
	for $\theta \bar q(0) \leq 1$, 
	then, for $\eta T/ \theta \geq 1$,
	\[
\bar q(T)
\leq 
\frac{1}{\eta T} \log\left(e+ \log\left( \frac{\eta T}{\theta}\right)\right)
	\]
\end{lemma}
\Beginproof
Firstly it is clear that $\bar q(t)$ is a decreasing sequence. 
Notice that the inequality \eqref{qbounds_III} rearranges to give 
\[
\frac{\log (e - \log ( \theta \bar q(t)) ))}{\bar q(t)^2}
\big[ 
\bar q(t) 
-
\bar q(t+1)
\big]
\geq \eta \, .
\]
Summing we see that 
\begin{equation}
\label{RS:sum_III}
\sum_{t=0}^{T-1}
\frac{\log (e - \log ( \theta \bar q(t)) )}{\bar q(t)^2}
\big[ 
\bar q(t) 
-
\bar q(t+1)
\big]
\geq
\eta T \, .	
\end{equation}
We could interpret this as a Riemann-Stieltjes approximation to the integral
\[
\int_{\bar q(T)}^{\bar q(0)}
\frac{\log (e - \log ( \theta \bar q) )}{\bar q^2}
d \bar q
\leq 
\frac{1}{\bar q(T)}
\log 
\left(
e+
\log \left(
\frac{1}{\theta \bar q(T)}\right)\right)
\]
In the inequality above we make the substitution $x=e^e \theta^{-1}\bar q^{-1}$ to the integral and we apply integral identity $\int \log \log x dx = x \log \log x - Li(x)$ and then note that the logarithmic integral is positive.

Since the function $\bar q \mapsto 	\frac{ \log(e-\log (\theta \bar q))}{\bar q^2}$ is decreasing on $(0,1/\theta)$ (which is easily verified by differentiation) the integral above upper bounds the sum in \eqref{RS:sum_III}, we have that 
\[
\frac{1}{\theta \bar q(T)}
\log\left(e+ \log\left( \frac{1}{\theta \bar q(T)}\right)\right)
\geq 
\frac{\eta T}{\theta}\, .
\]
Lemma \ref{new_lemma} given below generalizes an inequality for Lambert's W-function. Applied to the bound above this Lemma \ref{new_lemma} proves that
\[
\theta \bar q(T)
\leq 
\frac{\theta}{T\eta} \log\left(e+ \log\left( \frac{\eta T}{\theta}\right)\right)
\]
as required.
\Endproof

The following lemma converts a Lyapunov bound into a rate of convergence for probabilities. It is analogous to the lower-bound on Lambert's W-function.

\begin{lemma}\label{new_lemma}
	If $F: \mathbb R_+ \rightarrow \mathbb R_+$ is an increasing function and $T$ is such that $F(T)\geq 1$ then
	\[
	\frac{1}{q} F\left(\frac{1}{q}\right) \geq T
	\]
	implies
	\[
	q \leq \frac{F(T)}{T}
	\]
\end{lemma}
\Beginproof
Let $\tilde q = F(T)/T$ then notice 
\[
\frac{1}{\tilde q} F\left(\frac{1}{\tilde q} \right)
=
\frac{T}{F(T)} F \left( \frac{T}{F(T)} \right)
\leq 
\frac{T}{F(T)} F(T) = T
\leq 
\frac{1}{q} F\left(\frac{1}{q}\right)\, .
\]
Since $F(x)$ is positive and increasing so is $x F(x)$. Thus the above bound implies
\[
\frac{1}{\tilde q} \leq \frac{1}{q} \, .
\]
Thus $q\leq \tilde{q}= F(T)/T$.
\Endproof

We replace Proposition \ref{first_lemma} with Proposition \ref{2nd_lemma_III}. The proof is essentially the same as Proposition \ref{first_lemma}, but a number of calculations are more technically involved.

\begin{proposition}\label{2nd_lemma_III}
	For $\alpha(p) = \frac{\beta}{\log(e-\log p)}$,
	there exists a positive constant $\hat x$ such that 
	\[
	\sup_{\bm p \in E(\hat x)}
	\mathbb E \big[ \tau(\hat x) 
	|
	\bm p (0) = \bm p
	\big]
	< \infty \, .
	\]
\end{proposition}
\Beginproof
We will show that 
\[
\frac{1}{p_{a^\star}(t)}
\log\left( e +
\log \left( \frac{1}{p_{a^\star}(t) } \right)
\right)
-
ct
\]
is a supermartingale for some $c>0$. 

Note that (as in the proof of Proposition \ref{first_lemma}) if we let $x(t) = p_{a^\star}(t)^{-1}$ then, a short calculation gives, 
\[
x(t+1) 
=
\begin{cases}
\frac{1}{1+\alpha}
x(t)
&
\text{ w.p. } 
\frac{r^\star}{x(t)}\, ,
\\
x(t)
\frac{	
	p_\star (t) x (t)
}{
	p_\star (t) x(t) - \alpha  
}
& \text{ w.p. } p_{\star}(t) r_{\star}(t)\, ,
\\
x(t) & \text{ otherwise.}
\end{cases}
\]
Where $r_{\star}(t)$ and $p_{\star}(t)$ are the reward and probability of playing the leading arm at time $t$.
We apply the shorthand 
$\alpha = \alpha(x(t)^{-1})$, $x=x(t)$, $r_{\star}=r_{\star}(t)$ and $p_{\star}=p_{\star}(t)$. We have that
\begin{subequations}
	\begin{align}
	& \mathbb E 
	\Big[
	x(t+1)\log\left( e +
	\log \left( x(t+1) \right)\right)
	\big| H(t)
	\Big]
	- x(t) \log\left( e +
	\log \left( x(t)\right)\right)
	\notag
	\\
	=
	&
	\frac{r^\star}{x}
	\Big[
	\frac{x}{1+\alpha}
	\log\left( e +
	\log \left( \frac{x}{1+\alpha}\right)\right)
	-x\log\left( e +
	\log \left( x \right)\right)
	\Big]
	\notag
	\\
	&
	+
	p_{\star} r_{\star} 
	\Big[
	x \cdot 
	\frac{p_{\star} x}{p_{\star} x - \alpha}
	\cdot 
	\log\left( e +
	\log \left(
	x \cdot 
	\frac{p_{\star} x}{p_{\star} x - \alpha}
	\right)\right)
	- 
	x \log\left( e +
	\log \left( x\right)\right)
	\Big]
	\notag
	\\
	=
	& 
	\frac{r^\star}{x}
	\Big[
	\frac{x}{1+\alpha}
	\log\left( e +
	\log \left( \frac{x}{1+\alpha}\right)\right)
	-x\log\left( e +
	\log \left( x \right)\right)
	\Big]
	\label{term1_III}
	\\
	&
	+
p_{\star} r_{\star} 
x \cdot 
\frac{\alpha}{p_{\star} x - \alpha}
\cdot 
\log\left( e +
\log \left(
x
\right)\right)
	\label{term2_III}
	\\
	&
	+
	p_\star r_\star x 
	\cdot 
		\frac{p_{\star} x}{p_{\star} x - \alpha}
	\cdot 
	\Big[
	\log\left( e +
	\log \left(
	x \cdot 
	\frac{p_{\star} x}{p_{\star} x - \alpha}
	\right)\right)
	- 
 \log\left( e +
	\log \left( x\right)\right)
	\Big]
	\label{term3_III}
	\end{align}
\end{subequations}
We upper-bound each of the three terms above. First \eqref{term1_III} can be bounded as follows,
\begin{align}
	&
	\frac{r^\star}{x}
	\Big[
	\frac{x}{1+\alpha}
	\log\left( e +
	\log \left( \frac{x}{1+\alpha}\right)\right)
	-x\log\left( e +
	\log \left( x \right)\right)
	\Big]
	\notag
	\\
	\leq
	& 
	\frac{r^\star}{x}
	\Big[
	\frac{x}{1+\alpha}
	\log\left( e +
	\log \left( {x}\right)\right)
	-x\log\left( e +
	\log \left( x \right)\right)
	\Big]
	\notag
	\\
	=
	&
	- r^\star \frac{\alpha}{1+\alpha} \log \left( e + \log x \right)
	\notag
	\\
	\leq 
	&
 	- r^\star  \alpha ( 1- \alpha) \log (e+\log(x))
 	\notag
 	\\
 	=
 	&
 	- \beta r^\star + \frac{r^\star \beta^2 }{\log(e+\log(x))}
 	\label{term1-done_III}
\end{align}
\noindent Second \eqref{term2_III} can be expanded as follows
\begin{align}
&
p_{\star} r_{\star} 
x \cdot 
\frac{\alpha}{p_{\star} x - \alpha}
\cdot 
\log\left( e +
\log \left(
x 
\right)\right)
\notag
\\
=
&
 \alpha r_{\star} 
 \cdot 
\frac{p_{\star} x}{p_{\star} x - \alpha}
\cdot 
\log\left( e +
\log \left(
x 
\right)\right)
\notag
\\
=
& \alpha r_{\star} 
\cdot 
\log\left( e +
\log \left(
x
\right)\right)
+
\alpha^2 r_{\star} 
\cdot 
\frac{1}{p_{\star} x - \alpha}
\cdot 
\log\left( e +
\log \left(
x
\right)\right)
\notag
\\
=
&
\beta r_\star 
+
\frac{\beta^2 r_\star 
}{
p_\star x - \alpha 
}
\cdot 
\frac{1}{\log(e+\log x)}
\label{term2-done_III}
\end{align}

Before considering \eqref{term3_III}, we now take a few moments to show that 
\begin{equation}\label{logbd}
  -\log (1-z)/(1-z) \leq z+ 4z^2\qquad\text{ for }\qquad 0\leq z\leq \frac{1}{3}
\end{equation}
We require
		\[
		-\log (1-z) \leq (1-z) ( z+4z^2) = z+ 3z^2-4z^3
		\]
		Both side agree at $z=0$ and differentiating gives a sufficient condition: that for $0 \leq z \leq 1/3$
		\[
		\frac{1}{1-z} 
		\leq 1+6 z - 12 z^2,
		\quad 
		\text{or equivalently}
		\quad 
		0 \leq 5 - 18z + 12z^2.
		\]
		Inspecting the quadratic term on the right hand side. It is positive for $z=1/3$ negative for $z=1$ and positive for $z=2$. 

Third \eqref{term3_III} obeys the follow sequence of inequalities,	
\begin{align}
&	p_\star r_\star x 
\cdot 
\frac{p_{\star} x}{p_{\star} x - \alpha}
\cdot 
\Big[
\log\left( e +
\log \left(
x \cdot 
\frac{p_{\star} x}{p_{\star} x - \alpha}
\right)\right)
- 
\log\left( e +
\log \left( x\right)\right)
\Big]
\notag
\\
=
&
-p_\star r_\star x 
\frac{\log( 1 - \alpha/p_\star x) }{1 - \alpha/p_\star x}
\frac{
	\Big[
	\log\left( e +
	\log \left(
	x \cdot 
	\frac{p_{\star} x}{p_{\star} x - \alpha}
	\right)\right)
	- 
	\log\left( e +
	\log \left( x\right)\right)
	\Big]
}{
	e +\log \left(
	x \cdot 
	\frac{p_{\star} x}{p_{\star} x - \alpha}
\right)
-
 e -
\log \left( x\right)
}
\notag
\\
\leq 
&
-p_\star r_\star x 
\frac{\log( 1 - \alpha/p_\star x) }{1 - \alpha/p_\star x}
\notag
	\\
\leq 
&
p_\star r_\star x \cdot 
\left[
\frac{\alpha}{p_\star x}
+4
\frac{\alpha^2}{p^2_\star x^2}
\right]
\notag
\\
\leq 
&
\frac{\beta}{\log(e+\log x)} 
+
\frac{2N\beta^2}{x (\log(e+\log x))^2} \, .
\label{term3-done_III}
\end{align}
In the first equality above, note the numerator of the first fraction cancels with the denominator of the second fraction. In first inequality above, we note that $(\log(b) - \log(a) )/(b-a) \leq 1$ for $b,a\geq 1$. The second inequality we apply \eqref{logbd}, and, in final inequality, we note that $p_{\star} \geq \frac{1}{N}$.

Applying bounds 
\eqref{term1-done_III}, 
\eqref{term2-done_III}
and 
\eqref{term3-done_III}
to 
\eqref{term1_III}, \eqref{term2_III} and
\eqref{term3_III} gives that
\begin{align*}
& \mathbb E 
\Big[
x(t+1)\log\left( e +
\log \left( x(t+1) \right)\right)
\big| H(t)
\Big]
- x(t) \log\left( e +
\log \left( x(t)\right)\right)
\notag
\\
\leq &
-  \beta r^\star 
+
\beta r_\star  
+ \frac{r^\star \beta^2 }{\log(e+\log(x))}
+
\frac{\beta^2 r_\star 
}{
	p_\star x - \alpha 
}
\cdot 
\frac{1}{\log(e+\log x)}
+
\frac{\beta}{\log(e+\log x)} 
+
\frac{2N\beta^2}{x (\log(e+\log x))^2} 
\\
= 
&
- \beta \Delta
+
O\Big({1}/{\log(e+\log x)}  \Big)\, .
\end{align*}
Thus for $\epsilon \in (0,1)$ there exists an $\hat x$ such that for all $x(t) \geq \hat x$ we have that 
\[
\mathbb E 
\Big[
x(t+1)\log\left( e +
\log \left( x(t+1) \right)\right)
\big| H(t)
\Big]
- x(t) \log\left( e +
\log \left( x(t)\right)\right)
\leq - \beta \Delta (1-\epsilon) =: -c
\]
The remainder of the proof is identical to the stopping time argument in Proposition \ref{first_lemma}. 
\Endproof

As a consequence the following holds.
\begin{corollary}
	\label{prop1_III}
	For $\alpha>0$ such that \eqref{alpha_cond} holds,
	there exists a positive constant $\hat x$ such that 
	\[
	\sup_{\bm p \in E(2)}
	\mathbb E \big[ \tau(2) 
	|
	\bm p(0) = \bm p
	\big]
	< \infty \, .
	\]	
\end{corollary}
We refer the reader to the proof in Section \ref{corol5}. The proof follows as given there.

Lemma \ref{gamma_bound_III} establishes show that $\gamma(p)$ is increasing and convex. The proof is elementary calculus. 

\begin{lemma}\label{gamma_bound_III}
	The step size function 
	\[
	\gamma(p)= \frac{p^2}{\log(e-\log p)}
	\]
	is increasing and convex on the interval $p\in (0,1)$.
\end{lemma}
\Beginproof
This can be proven by differentiating $\gamma(p)$.
\begin{align*}
\gamma'(p) 
&
= 
\frac{2p}{\log(e-\log p)}
+ 
\frac{p}{(e-\log p)\log(e-\log p)^2}
 >0
\end{align*}
So $\gamma(p)$ is increasing.
Differentiating once more,
\begin{align*}
\gamma''(p)
&
=
\frac{2}{\log(e-\log p)}
+
\frac{2}{(e-\log p)\log(e-\log p)^2}
+
\frac{1}{(e-\log p)\log(e-\log p)^2}
\\
&
+
\frac{1}{(e-\log p)^2\log(e-\log p)^2}
+
\frac{2}{p(e-\log p)^2\log(e-\log p)^3}
\\
&> 0
\end{align*}
So the function $\gamma(p)$ is increasing and convex.
\Endproof

The following result replaces Proposition \ref{LemExpectq}. 

\begin{proposition}
	\label{LemExpectq_2_III}
	For the process $(\hat q(s) : s\in \mathbb Z_+)$ \\
	a)
	\begin{equation}\label{qIneq_III}
	\mathbb E [ \hat q(s+1) | \hat H(s)  ]
	-
	\hat q(s) 
	\leq 
	-\Delta \beta
	\frac{1}{N}
	\frac{\hat q(s)^2}{
		\log(e-\log (\frac{\hat q(s)}{N}))} \, .
	\end{equation}
	and, thus, it is a positive supermartingale.\\
	b) With probability $1$,
	\[
	\hat q(s) \xrightarrow[s\rightarrow \infty]{} 0\, .
	\]
	c)  For $s\beta \Delta \geq e$,
	\begin{equation}\label{qTime2_III}
	\mathbb E [ \hat q (s) ]
	\leq 
	\frac{N}{ \beta \Delta T} \log\left(e+ \log\left( {\beta \Delta T}\right)\right)  .	
	\end{equation}
\end{proposition}
\Beginproof 
	a)  By the identical argument used to derive \eqref{q4} in Proposition \ref{LemExpectq}, we have that
	\begin{align}
	\mathbb E [ \hat q(s+1) | \hat H(s)  ]
	-
	\hat q(s) 
	&
	\leq 
	- \Delta\!\!\! \sum_{a: a\neq a^\star} 
	\alpha (p_a(t_s) )
	p_a(t_s)^2
	=
	- \Delta\!\!\! \sum_{a: a\neq a^\star} 
	\beta
	\frac{p_a(t_s)^2}{\log(e-\log (p_a(t_s)))} \, .
	\label{q4_2}
	\end{align}
	By Lemma \ref{gamma_bound_III} the terms in the righthand side of \eqref{q4_2} are convex. So
	by Jensen's inequality:
	\[
	\frac{1}{N-1}
	\sum_{a: a\neq a^\star} 
	\frac{p_a^2}{\log(e-\log (p_a(t_s)))}
	\geq 
	\frac{
		\left( 
		\sum_{a\neq a^\star} 
		\frac{p_a }{N-1}
		\right)^2
	}{
		\log\left(e-\log \left(
		\sum_{a\neq a^\star}
		\frac{p_a }{N-1}
		\right)\right)
	}
	=
	\frac{1}{(N-1)^2}
	\frac{\hat q^2}{
		\log (e-\log(\frac{\hat q}{N-1} )) }\, .
	\]
	Applying this to \eqref{q4_2} gives
	\[
	\mathbb E [ \hat q(s+1) | \hat H(s)  ]
	-
	\hat q(s) 
	\leq 
	-\Delta \beta
	\frac{1}{(N-1)}
	\frac{\hat q^2}{
		\log (e-\log(\frac{\hat q}{N-1} ))} 
	\leq 
	-\Delta \beta
	\frac{1}{N}
	\frac{\hat q^2}{
		\log (e-\log(\frac{\hat q}{N} ))} \, .
	\]
	Thus \eqref{qIneq} holds as required.
	
	b) The proof of this part is identical to the argument in Proposition \ref{LemExpectq}. So we do not repeat it here.
	
	c) Taking expectations on both sides of \eqref{qIneq} and applying Jensen's Inequality again,
	\begin{align*}
	\mathbb E [ \hat q (s+1) ]
	-
	\mathbb E[\hat q (s) ] 
	&
	\leq 
	-\beta N \Delta 
	\mathbb E 
	\left[
	\frac{ (\frac{\hat q(s)}{N})^2}{
		\log (e-\log(\frac{\hat q(s)}{N} ))} 
	\right]
	\\
	&
	\leq 
	-\beta N \Delta 
	\frac{
		\left(\frac{\mathbb E[\hat q(s)]}{N}\right)^2 }{
\log (e-\log(\frac{\mathbb E [\hat q(s)]}{N} ))
	}
	=
	-\frac{
		\beta \Delta}{N}
	\frac{\mathbb E[ \hat q (s) ]^2
	}{
\log (e-\log(\frac{\mathbb E [\hat q(s)]}{N} )
	}	\, .
	\end{align*}

	Thus applying Lemma \ref{upperbound_2_III} (with $\eta = \beta \Delta /N$ and $\theta = 1/N$), we have that 
	\[
	\mathbb E [\hat q(s) ]
	\leq 
	\frac{N  }{\beta \Delta s}
\log\left(e+
	\log \Big(
	\frac{\beta\Delta s}{ N \frac{1}{N}} 
	\Big) \right)
	=\frac{N}{ \beta \Delta T} \log\left(e+ \log\left( {\beta \Delta T}\right)\right) \, ,
	\]
as required.
\Endproof

We state Proposition \ref{trans_prob_prop_2_III} below. The result is a restatement of Proposition \ref{trans_prob_prop}. 

\begin{proposition}
	\label{trans_prob_prop_2_III}
	a) If $q_{a^\star}(0) \leq  \frac{1}{2}$ then, there exists a constant $\rho < 1$ such that 
	\begin{equation}\label{hit_prob_1_III}
	\mathbb P ( \sigma^{(1)} < \infty ) < \rho\, .		
	\end{equation}
	Moreover
	\begin{equation}\label{hit_prob_2_III}
	\mathbb P ( \sigma^{(k)} < \infty | \sigma^{(k-1)} <\infty)
	< \rho \, .
	\end{equation}
	b) 
	\[
	\sum_{t=0}^\infty
	\mathbb P 
	\Big(
	q_{a^\star} (t) \geq \frac{1}{2}
	\Big)	
	< \infty \, .
	\]
	c) With probability $1$,
	\[
	q_{a^\star}(t) \xrightarrow[t\rightarrow \infty ]{}
	0.
	\]
\end{proposition}

\noindent The proof of Proposition \ref{trans_prob_prop_2_III}  is identical to the proof of Proposition \ref{trans_prob_prop}, with the only change being which results are applied: we replace Proposition   \ref{LemExpectq} (for the supermartingale property) with Proposition \ref{LemExpectq_2_III}; Corollary \ref{prop1} with Corollary \ref{prop1_III}; Proposition \ref{LemExpectq}b) with Proposition \ref{LemExpectq_2_III}b). We refer the reader to the proof of Proposition \ref{trans_prob_prop}.

We can upper bound the sum of error terms as follows.

\begin{lemma}\label{last_lemma_III}
	For $\theta>0$,
	\[
	\sum_{t=2}^T
	\frac{\log (e+\log \theta t)}{ t}
	\leq 
	\log(T)
	\cdot \log (e+\log (\theta T))\, .
	\]
\end{lemma}
\Beginproof
\[
\sum_{t=2}^T
\frac{\log (e+\log \theta t)}{ t}
\leq 
{\log (e+\log \theta T)}
\sum_{t=2}^T
\frac{1}{t}
\leq 
\log (T)
\cdot  
\log(e+\log (\theta T))
\]
\Endproof

We can now prove Theorem \ref{3rd_theorem}. 

\begin{proof}[Proof of Theorem \ref{3rd_theorem}.]
Note that by Proposition \ref{trans_prob_prop_2_III}c), we have that
\[
p_{a^\star}(t) = 1- q_{a^\star}(t) \xrightarrow[t\rightarrow\infty]{} 1.
\]
By the identical argument applied in Theorem \ref{main_theorem} to bound the regret, we have that
\[
\mathcal R\! g(T)
\leq 
Q
+
\sum_{s=0}^{T-1} 
\mathbb E \big[ 
\hat q(s) 
\big]\, .
\]
Applying Proposition \ref{LemExpectq_2_III} 
and Lemma \ref{last_lemma_III} 
\begin{align*}
\sum_{s=0}^{T-1} 
\mathbb E \big[ 
\hat q(s) 
\big]
&
\leq 
1
+
\sum_{s=2 }^{T} 
\mathbb E \big[ 
\hat q(s) 
\big]	
\\	
&
\leq 
1
+
\sum_{s=2 }^{T} 	
\frac{N}{\beta \Delta s} \log\left(e+ \log\left( {\beta \Delta s}\right)\right) 
\\
&
\leq 
1
+
\frac{N}{\beta\Delta}
	\log(T)
\cdot \log (e+\log (\beta \Delta T))\, .
\end{align*}
Combining this with the above inequality (and noting that $\beta \Delta \leq 1$) gives 
\begin{equation*}
\mathcal R\! g(T)
\leq 
\frac{N}{\beta\Delta}
\log T
\cdot \log (e+\log ( T))
+
Q
+
1 
\end{equation*}
as required.
\end{proof}

%% file: THEOREM_3e.tex
\section{Theorem \ref{4th_theorem}.}\label{sec:4th_theorem}
We note that the extra logarithmic factor in Theorem \ref{3rd_theorem} can be improved further by dividing the learning rate by a more slowly increasing function.  This is proven in the following theorem. 

\begin{theorem}
	\label{4th_theorem}
	Let $\alpha: [0,1] \rightarrow [0,1]$ be such that
	\begin{equation}\label{alpha_2nd_cond_IIIe}
	\alpha(p) = 
	\frac{1}{p^2}\int_0^p\int_0^v l(u)du dv,	
	\end{equation}
	where $l(u)\in [0,1]$ is a monotone, increasing, slowly varying function such that 
	$\lim_{u\to 0}l(u)=0$. 
	Then the SAMBA process $(p_a(t) : a\in\mathcal A)$, $t\in \mathbb Z_+$, is a Markov chain such that, with probability $1$, 
	$
	p_{a^\star} (t) \rightarrow {} 1
	$ as $t\rightarrow\infty$
	and
	\begin{equation*}
	\mathcal R\! g(T)
	\leq 
	\frac{CN}{\Delta l\left( \frac{C}{\Delta T}\right)}\cdot \log (T) +
	Q
	+
	1 
	\end{equation*}
	for some $C>0$. 
\end{theorem}
The definition and examples of slowly varying functions are given in Appendix~\ref{sec:rv}.

The proof follows a similar pattern to Theorem \ref{3rd_theorem}. 
Some results are identical. 
These results we state but refer the reader to the earlier proof. The main changes required are 
to replace Lemma \ref{upperbound_2_III} with Lemma \ref{upperbound_2_IIIe}, Proposition \ref{2nd_lemma_III} with Proposition \ref{2nd_lemma_IIIe}, 
Lemma \ref{gamma_bound_III} with Lemma \ref{gamma_bound_IIIe}. 
We state and prove these results. 
We also restate results whose proof is unchanged and refer the reader to the appropriate results in the main body of the paper.

First note that $\gamma(p)$ has the following form,
\begin{equation}\label{eq:general.gamma}
    \gamma(p) = \int_0^p \int_0^u l(t)dtdu.
\end{equation}
Since $l$ is slowly varying function, applying the Karmata theorem (see part (ii) of Theorem~\ref{thm:karamata} below) 
two times we can observe that $\gamma$ is regularly varying at $0$  of index $2$.

Next we can immediately see that  
\begin{lemma}\label{gamma_bound_IIIe}
	The step size function 
		$\gamma(p)$ 	
    is increasing and convex on the interval $p\in (0,1)$.
\end{lemma}
\Beginproof
The statement  follows  from~\eqref{eq:general.gamma} and the fact that 
$\gamma''(p)=l(p)\ge 0$ for $p\in(0,1)$. 
\Endproof

\medskip
\noindent We can then analyse the discrete analogue of the o.d.e. $\dot q= - \eta \gamma(q)$.


\begin{lemma}
	\label{upperbound_2_IIIe}
	If $\bar q(t)$ is a sequence of real numbers belonging to the interval $(0,1)$ such that 
	\begin{equation}\label{qbounds_IIIe}
	\bar q (t+1)
	-
	\bar q (t)
	\leq 
	-\eta \gamma (\theta \bar q(t))
	\end{equation}
	for $\theta \bar q(0) \leq 1$.
	Then, for monotone increasing function $l$, 
	there exists a  constant $C$ such that for 
	$l\left(\frac{C}{T \eta \theta}\right) \le 1$,
	\[
\bar q(T)
\leq 
\frac{C}{T\eta \theta^2} \frac{1}{l\left( \frac{C}{\eta \theta T}\right)}. 
	\]
\end{lemma}
\Beginproof
Firstly it is clear that $\bar q(t)$ is a decreasing sequence. 
Notice that the inequality \eqref{qbounds_IIIe} rearranges to give 
\[
\frac{\bar q(t) 
-
\bar q(t+1)
}{\gamma (\theta \bar q(t))}
\geq \eta \, .
\]
Summing we see that 
\begin{equation}
\label{RS:sum_IIIe}
\sum_{t=0}^{T-1}
\frac{1}{\gamma (\theta \bar q(t))}
\big[ 
\bar q(t) 
-
\bar q(t+1)
\big]
\geq
\eta T \, .	
\end{equation}
We could interpret this as a Riemann-Stieltjes approximation to the integral
\[
\int_{\bar q(T)}^{\bar q(0)}
\frac{1}{\gamma (\theta \bar q)}
d \bar q
\leq 
\frac{C\bar q(T)}{\gamma(\theta\bar q(T))}
\]
for some constant  $C$ depending on $\gamma$. 
The latter inequality follows from the Karamata theorem for regularly varying functions, 
see part (i) of Theorem~\ref{thm:karamata} below. 

The integral above upper bounds the sum in \eqref{RS:sum_IIIe}, we have that 
\[
	\frac{C\bar q(T)}{\gamma(\theta\bar q(T))}\geq 
\eta T\, .
\]
Note that applying  Karamata theorem (see part (ii) of Theorem~\ref{thm:karamata} below) two times 
we obtain  $\alpha(p)\sim \frac{1}{2}l(p), p\to 0$. 
Hence, increasing $C$ if necessary,  we obtain 
\[
	\frac{C}{\theta^2\bar q(T)l(\theta\bar q(T))}\geq 
\eta T\, 
\] 
We can apply Lemma~\ref{new_lemma} now to obtain 
\[
\bar q(T)
\leq 
\frac{C}{T\eta \theta^2} \frac{1}{l\left( \frac{C}{\eta \theta T}\right)}
\]
as required.
\Endproof

We replace Proposition \ref{2nd_lemma_III} with Proposition \ref{2nd_lemma_IIIe}. The statement is similar to Proposition \ref{first_lemma}. The proof is essentially the same as Proposition \ref{2nd_lemma_III}, but a number of calculations are more technically involved.

\begin{proposition}\label{2nd_lemma_IIIe}
	There exists a positive constant $\hat x$ such that 
	\[
	\sup_{\bm p \in E(\hat x)}
	\mathbb E \big[ \tau(\hat x) 
	|
	\bm p (0) = \bm p
	\big]
	< \infty \, .
	\]
\end{proposition}
\Beginproof
For $x\ge 1$, let 
\[
    f(x) = \int_1^x \frac{dt}{\alpha(1/t)}
\]
We will show that 
\[
f\left(\frac{1}{p_{a^\star}(t)}\right)
-
ct
\]
is a supermartingale for some $c>0$. 

Note that (as in the proof of Proposition \ref{first_lemma}) if we let $x(t) = p_{a^\star}(t)^{-1}$ then, a short calculation gives, 
\[
x(t+1) 
=
\begin{cases}
\frac{1}{1+\alpha}
x(t)
&
\text{ w.p. } 
\frac{r^\star}{x(t)}\, ,
\\
x(t)
\frac{	
	p_\star (t) x (t)
}{
	p_\star (t) x(t) - \alpha  
}
& \text{ w.p. } p_{\star}(t) r_{\star}(t)\, ,
\\
x(t) & \text{ otherwise.}
\end{cases}
\]
Where $r_{\star}(t)$ and $p_{\star}(t)$ are the reward and probability of playing the leading arm at time $t$.
We apply the shorthand 
$\alpha = \alpha(x(t)^{-1})$, $x=x(t)$, $r_{\star}=r_{\star}(t)$ and $p_{\star}=p_{\star}(t)$. We have that
\begin{subequations}
	\begin{align}
	\mathbb E 
	\left[
	f(x(t+1))
	\mid H(t)
	\right]
	- f(x(t)) 
		=
	&
	\frac{r^\star}{x}
    \left[	
        f\left(
            \frac{x}{1+\alpha}
            \right) -f(x)
	\right ]
	\label{term1_IIIe}
	\\
	&
	+
	p_{\star} r_{\star} 
	\left[
	f\left(x \cdot 
    \frac{p_{\star} x}{p_{\star} x - \alpha}
    \right)
	- 
    f(x)
    \right]
	\label{term2_IIIe}
	\end{align}
\end{subequations}
We upper-bound each of the  terms above. 
First \eqref{term1_IIIe} can be bounded as follows,
\begin{align*}
    \frac{r^\star}{x}
    \left[	 
        f\left(
            \frac{x}{1+\alpha}
            \right) -f(x)
	\right ]
    &
    =
    \frac{r^\star}{x}
    \int_x^{\frac{x}{1+\alpha}}
    \frac{dt}{\alpha(1/t)}
	\\
    &\le
	-r^\star \frac{\alpha}{1+\alpha} 
	\frac{1}{\max_{y\in\left[\frac{x}{1+\alpha},x\right]}\alpha(1/y)}.    
\end{align*}
Now, by the Karamata theorem (see part (ii) of Theorem~\ref{thm:karamata} below), 
$
	\alpha(p) =\frac{\gamma(p)}{p^2} \sim \frac{1}{2}l(p)\to 0, \text{ as } p\to 0,
$
which (for $x=p^{-1}$) implies that 
\[
	\frac{1}{1+\alpha}\to 1, \quad x\to\infty. 
\]
Also, since $\alpha$ is slowly varying at $0$,  
if follows from the Uniform Convergence Theorem for slowly varying functions 
(see Theorem~\ref{thm:uct} below) that 
\[
	\frac{\alpha(1/x)}{\max_{y\in\left[\frac{x}{1+\alpha},x\right]}\alpha(1/y)}\to 1, \quad x\to\infty.
\]
As a result, there exists $\widehat x$ such that for $x>\widehat x$
\begin{equation}
	\frac{r^\star}{x}
    \left[	
        f\left(
            \frac{x}{1+\alpha}
            \right) -f(x)
	\right ]
 	\le -r^\star +\frac{\Delta}{2}. 
 	\label{term1-done_IIIe}
\end{equation}
\noindent Second \eqref{term2_IIIe} can be estimated as follows
\begin{align*}
	p_{\star} r_{\star} 
	\left[
	f\left(x \cdot 
    \frac{p_{\star} x}{p_{\star} x - \alpha}
    \right)
	- 
    f(x)
	\right]
	&= p_{\star} r_{\star}  \int_x^{x \cdot 
	\frac{p_{\star} x}{p_{\star} x - \alpha}} \frac{dt}{\alpha(1/t)}\\
	&\le 
	p_{\star} r_{\star}
	\frac{\alpha x}{p_{\star} x - \alpha}
	\frac{1}{
		\min_{t\in [x, x  \frac{p_{\star} x}{p_{\star} x - \alpha}]}\alpha(1/t)}		
\end{align*}	
As earlier, it follows from slow variation of $\alpha$  that 
\[
	\frac{\alpha(1/x)}{
		\min_{t\in [x, x  \frac{p_{\star} x}{p_{\star} x - \alpha}]}\alpha(1/t)}		
		\to 1, \quad x\to\infty
\]
Also, since $\alpha(1/x)\to 0$  as $x\to \infty$, 
\[
	\frac{p_{\star}  x}{p_{\star} x - \alpha} \to 1. 
\]
Hence, there exists $\widehat x$ such that for $x>\widehat x$, 
\begin{align}
	p_{\star} r_{\star} 
	\left[
	f\left(x \cdot 
    \frac{p_{\star} x}{p_{\star} x - \alpha}
    \right)
	- 
    f(x)
	\right]
	\le r_{\star} +\frac{\Delta}{4}. 
\label{term2-done_IIIe}
\end{align}
Applying bounds 
\eqref{term1-done_IIIe} 
and 
\eqref{term2-done_IIIe}
 gives that for $x>\widehat x$, 
\begin{align*}
	\mathbb E 
	\left[
	f(x(t+1))
	\mid H(t)
	\right]
	- f(x(t)) 
	\le -r^{\star}+r_\star+\frac{3\Delta}{4}\le -\frac{\Delta}{4}. 
\end{align*}
Thus we can put $c=\frac{\Delta}{4}$. 
The remainder of the proof is identical to the stopping time argument in Proposition \ref{first_lemma}.
\Endproof

As a consequence the following holds.
\begin{corollary}
	\label{prop1_IIIe}
	For $\alpha>0$ such that \eqref{alpha_cond} holds,
	there exists a positive constant $\hat x$ such that 
	\[
	\sup_{\bm p \in E(2)}
	\mathbb E \big[ \tau(2) 
	|
	\bm p(0) = \bm p
	\big]
	< \infty \, .
	\]	
\end{corollary}
We refer the reader to the proof in Section \ref{corol5}. The proof follows as given there.

The following result replaces Proposition \ref{LemExpectq_2_III}. 
The proof is done in exactly the same way using the Jensen inequality and convexity 
of $\gamma$. 
 We refer the reader to the proof of Proposition \ref{LemExpectq_2_III} for details.

\begin{proposition}
	\label{LemExpectq_2_IIIe}
	For the process $(\hat q(s) : s\in \mathbb Z_+)$ \\
	a)
	\begin{equation}\label{qIneq_IIIe}
	\mathbb E [ \hat q(s+1) | \hat H(s)  ]
	-
	\hat q(s) 
	\leq 
	-\Delta 
	(N-1)
	\gamma\left(
			\frac{\hat q(s)}{N-1}\right) \, , 
	\end{equation}
	and, thus, it is a positive supermartingale.\\
	b) With probability $1$,
	\[
	\hat q(s) \xrightarrow[s\rightarrow \infty]{} 0\, .
	\]
	c)  There exists a constant $C$ such that  
	\begin{equation}\label{qTime2_III}
	\mathbb E [ \hat q (s) ]
	\leq 
	\frac{CN}{  \Delta T}\frac{1}{l\left( \frac{C}{\Delta T}\right)} .	
	\end{equation}
\end{proposition}

We state Proposition \ref{trans_prob_prop_2_IIIe} below. The result is a restatement of Proposition \ref{trans_prob_prop}. The proof is identical to the argument there and we refer the reader to the proof of Proposition \ref{trans_prob_prop}.

\begin{proposition}
	\label{trans_prob_prop_2_IIIe}
	a) If $q_{a^\star}(0) \leq  \frac{1}{2}$ then, there exists a constant $\rho < 1$ such that 
	\begin{equation}\label{hit_prob_1_IIIe}
	\mathbb P ( \sigma^{(1)} < \infty ) < \rho\, .		
	\end{equation}
	Moreover
	\begin{equation}\label{hit_prob_2_IIIe}
	\mathbb P ( \sigma^{(k)} < \infty | \sigma^{(k-1)} <\infty)
	< \rho \, .
	\end{equation}
	b) 
	\[
	\sum_{t=0}^\infty
	\mathbb P 
	\Big(
	q_{a^\star} (t) \geq \frac{1}{2}
	\Big)	
	< \infty \, .
	\]
	c) With probability $1$,
	\[
	q_{a^\star}(t) \xrightarrow[t\rightarrow \infty ]{}
	0.
	\]
\end{proposition}



We can now prove Theorem \ref{3rd_theorem}. 

\proof{Proof of Theorem \ref{3rd_theorem}.}
Note that by Proposition \ref{trans_prob_prop_2_IIIe}c), we have that
\[
p_{a^\star}(t) = 1- q_{a^\star}(t) \xrightarrow[t\rightarrow\infty]{} 1.
\]
By the identical argument applied in Theorem \ref{main_theorem} to bound the regret, we have that
\[
\mathcal R\! g(T)
\leq 
Q
+
\sum_{s=0}^{T-1} 
\mathbb E \big[ 
\hat q(s) 
\big]\, .
\]
Applying Proposition \ref{LemExpectq_2_IIIe}, 
\begin{align*}
\sum_{s=0}^{T-1} 
\mathbb E \big[ 
\hat q(s) 
\big]
&
\leq 
1
+
\sum_{s=2 }^{T} 
\mathbb E \big[ 
\hat q(s) 
\big]	
\leq 
1
+
\sum_{s=2 }^{T} 
\frac{CN}{  \Delta s}\frac{1}{l\left( \frac{C}{\Delta s}\right)}	
\\
&
\leq 
1
+
\frac{CN}{\Delta l\left( \frac{C}{\Delta T}\right)}
\sum_{2}^T\frac{1}{s}	
\le 
1
+
\frac{CN}{\Delta l\left( \frac{C}{\Delta T}\right)} \log T
.
\end{align*}
Combining this with the above inequality 
\begin{equation*}
\mathcal R\! g(T)
\leq 
\frac{CN}{\Delta l\left( \frac{C}{\Delta T}\right)} \log T +
Q
+
1 
\end{equation*}
as required.
\hfill$\square$\endproof

%% file: REGULAR_VARIATION.tex
\section{Regular Variation.}\label{sec:rv}
In this section we will collect  information about regular variation. 
More detailed information can be found in~\cite[Chapter XIII.8-9]{F71} or~\cite{bingham_goldie_teugels_1987}. 
\begin{definition}
    A positive function $l$ defined on $[0,+\infty)$ varies 
    slowly at infinity (at $0$) is 
    \[
      \lim_{x\to \infty}\frac{l(tx)}{l(x)}  =1 
       \left(
          \mbox {respectively }
        \lim_{x\to 0}\frac{l(tx)}{l(x)}  =1 
      \right). 
    \]
    A function $U$ varies regularly  at infinity (at $0$)   
    of  index $\rho (-\infty<\rho<+\infty)$  if 
    $U(x)=x^\rho l(x)$, where $l$ is slowly varying at infinity (at $0$).
\end{definition}
Examples of slowly varying functions at infinity are 
$\log^c(1+x), \log^c (1+\log (1+x))$. Note that it follows from the definition 
that if $l(x)$ is slowly varying at infinity then $l(1/x)$ is 
slowly  varying at $0$. Thus,  $\log^c(1+1/x), \log (1+\log (1+1/x))$ 
are examples of functions slowly varying at $0$. 
The following version of  the Karamata theorem can be derived from  
\cite[Chapter XIII.9, Theorem 1]{F71} by  using the transformation $1/x$ 
that transforms function regularly varying at infinity to functions regularly varying at $0$. 
\begin{theorem}\label{thm:karamata}
    \begin{enumerate}[(i)]
    \item 
    Let $U$ be regularly varying at $0$ of index $\rho<-1$. Then, 
    \[
        \int_x^{x_0} U(t) dt \sim \frac{1}{-\rho-1}x U(x),\quad x\to 0,
    \]
    for any fixed $x_0>0$. 
    \item 
    Let $U$ be regularly varying at $0$ of index $\rho>-1$. Then, 
    \[
        \int_0^{x} U(t) dt \sim \frac{1}{\rho+1}x U(x),\quad x\to 0.
    \]
    \end{enumerate}
\end{theorem}
We will also need the Uniform Convergence Theorem for slowly varying functions,
see  Theorem 1.2.1 in~\cite{bingham_goldie_teugels_1987}.
\begin{theorem}\label{thm:uct}
    If $l$ is slowly varying at infinity (at $0$) then 
    \[
        \frac{l(\lambda x)}{l(x)} \to 1, \mbox {as } x\to \infty (\mbox{respectively } x\to 0)
    \]
    uniformly on each $\lambda$-compact set on $(0,\infty)$.
\end{theorem}

%% file: SAMBA-Arxiv.bbl
\begin{thebibliography}{37}
\providecommand{\natexlab}[1]{#1}
\providecommand{\url}[1]{\texttt{#1}}
\expandafter\ifx\csname urlstyle\endcsname\relax
  \providecommand{\doi}[1]{doi: #1}\else
  \providecommand{\doi}{doi: \begingroup \urlstyle{rm}\Url}\fi

\bibitem[Abernethy et~al.(2009)Abernethy, Hazan, and
  Rakhlin]{abernethy2009competing}
J.~D. Abernethy, E.~Hazan, and A.~Rakhlin.
\newblock Competing in the dark: An efficient algorithm for bandit linear
  optimization.
\newblock 2009.

\bibitem[Agarwal et~al.(2019)Agarwal, Kakade, Lee, and
  Mahajan]{agarwal2019optimality}
A.~Agarwal, S.~M. Kakade, J.~D. Lee, and G.~Mahajan.
\newblock Optimality and approximation with policy gradient methods in markov
  decision processes.
\newblock \emph{arXiv preprint arXiv:1908.00261}, 2019.

\bibitem[Agrawal(1995)]{agrawal_1995}
R.~Agrawal.
\newblock Sample mean based index policies by o(log n) regret for the
  multi-armed bandit problem.
\newblock \emph{Advances in Applied Probability}, 27\penalty0 (4):\penalty0
  1054–1078, 1995.
\newblock \doi{10.2307/1427934}.

\bibitem[Agrawal and Goyal(2012)]{Agrawal2012}
S.~Agrawal and N.~Goyal.
\newblock {Analysis of Thompson Sampling for the Multi-armed Bandit Problem}.
\newblock In \emph{25th Annual Conference on Learning Theory}, volume~23, pages
  39.1----39.26, 2012.

\bibitem[Anscombe(1963)]{anscombe1963sequential}
F.~Anscombe.
\newblock Sequential medical trials.
\newblock \emph{Journal of the American Statistical Association}, 58\penalty0
  (302):\penalty0 365--383, 1963.

\bibitem[Auer et~al.(1995)Auer, Cesa-Bianchi, Freund, and
  Schapire]{auer1995gambling}
P.~Auer, N.~Cesa-Bianchi, Y.~Freund, and R.~E. Schapire.
\newblock Gambling in a rigged casino: The adversarial multi-armed bandit
  problem.
\newblock In \emph{Proceedings of IEEE 36th Annual Foundations of Computer
  Science}, pages 322--331. IEEE, 1995.

\bibitem[Auer et~al.(2002)Auer, Cesa-Bianchi, and Fischer]{auer2002finite}
P.~Auer, N.~Cesa-Bianchi, and P.~Fischer.
\newblock Finite-time analysis of the multiarmed bandit problem.
\newblock \emph{Machine learning}, 47\penalty0 (2-3):\penalty0 235--256, 2002.

\bibitem[Babichev and Bach(2018)]{babichev2018constant}
D.~Babichev and F.~Bach.
\newblock Constant step size stochastic gradient descent for probabilistic
  modeling.
\newblock \emph{arXiv preprint arXiv:1804.05567}, 2018.

\bibitem[Beck and Srikant(2012)]{beck2012error}
C.~L. Beck and R.~Srikant.
\newblock Error bounds for constant step-size q-learning.
\newblock \emph{Systems \& Control Letters}, 61\penalty0 (12):\penalty0
  1203--1208, 2012.

\bibitem[Bertsimas and Tsitsiklis(1993)]{bertsimas1993}
D.~Bertsimas and J.~Tsitsiklis.
\newblock Simulated annealing.
\newblock \emph{Statist. Sci.}, 8\penalty0 (1):\penalty0 10--15, 02 1993.
\newblock \doi{10.1214/ss/1177011077}.

\bibitem[Bhandari and Russo(2019)]{bhandari2019global}
J.~Bhandari and D.~Russo.
\newblock Global optimality guarantees for policy gradient methods.
\newblock \emph{arXiv preprint arXiv:1906.01786}, 2019.

\bibitem[Bingham et~al.(1987)Bingham, Goldie, and
  Teugels]{bingham_goldie_teugels_1987}
N.~H. Bingham, C.~M. Goldie, and J.~L. Teugels.
\newblock \emph{Regular Variation}.
\newblock Encyclopedia of Mathematics and its Applications. Cambridge
  University Press, 1987.
\newblock \doi{10.1017/CBO9780511721434}.

\bibitem[Bouneffouf and Rish(2019)]{bouneffouf2019survey}
D.~Bouneffouf and I.~Rish.
\newblock A survey on practical applications of multi-armed and contextual
  bandits.
\newblock \emph{arXiv preprint arXiv:1904.10040}, 2019.

\bibitem[Bubeck and Cesa-Bianchi(2012)]{bubeck2012regret}
S.~Bubeck and N.~Cesa-Bianchi.
\newblock Regret analysis of stochastic and nonstochastic multi-armed bandit
  problems.
\newblock \emph{Foundations and Trends{\textregistered} in Machine Learning},
  5\penalty0 (1):\penalty0 1--122, 2012.

\bibitem[Denisov et~al.(2016)Denisov, Korshunov, and Wachtel]{denisov2016edge}
D.~Denisov, D.~Korshunov, and V.~Wachtel.
\newblock At the edge of criticality: Markov chains with asymptotically zero
  drift.
\newblock \emph{arXiv preprint arXiv:1612.01592}, 2016.

\bibitem[Denisov et~al.(2020)Denisov, Korshunov, and
  Wachtel]{denisov2020renewal}
D.~Denisov, D.~Korshunov, and V.~Wachtel.
\newblock Renewal theory for transient markov chains with asymptotically zero
  drift.
\newblock \emph{Trans. Amer. Math. Soc. (to appear)}, 2020.
\newblock \doi{https://doi.org/10.1090/tran/8167}.

\bibitem[Dieuleveut et~al.(2017)Dieuleveut, Durmus, and
  Bach]{dieuleveut2017bridging}
A.~Dieuleveut, A.~Durmus, and F.~Bach.
\newblock Bridging the gap between constant step size stochastic gradient
  descent and markov chains.
\newblock \emph{arXiv preprint arXiv:1707.06386}, 2017.

\bibitem[Dupuis and Williams(1994)]{dupuis1994lyapunov}
P.~Dupuis and R.~J. Williams.
\newblock Lyapunov functions for semimartingale reflecting brownian motions.
\newblock \emph{The Annals of Probability}, 22\penalty0 (2):\penalty0 680--702,
  1994.

\bibitem[Feller(1971)]{F71}
W.~Feller.
\newblock \emph{An Introduction to Probability Theory and its Applications,
  volume 2}.
\newblock Wiley mathematical statistics series. Willey, New
  York-London-Sydney-Toronto, 1971.

\bibitem[Gittins et~al.(2011)Gittins, Glazebrook, and Weber]{gittins2011multi}
J.~Gittins, K.~Glazebrook, and R.~Weber.
\newblock \emph{Multi-armed Bandit Allocation Indices}.
\newblock Wiley, 2011.
\newblock ISBN 9781119990215.

\bibitem[Hajek(1986)]{hajek}
B.~Hajek.
\newblock \emph{Optimization by simulated annealing: a necessary and sufficient
  condition for convergence}, volume~8 of \emph{Lecture Notes--Monograph
  Series}, pages 417--427.
\newblock Institute of Mathematical Statistics, Hayward, CA, 1986.
\newblock \doi{10.1214/lnms/1215540316}.

\bibitem[Hazan(2016)]{hazan2016introduction}
E.~Hazan.
\newblock Introduction to online convex optimization.
\newblock \emph{Foundations and Trends in Optimization}, 2\penalty0
  (3-4):\penalty0 157--325, 2016.

\bibitem[Kaufmann and Korda(2012)]{Kaufmann2012}
E.~Kaufmann and N.~Korda.
\newblock {Thompson Sampling : An Asymptotically Optimal Finite Time Analysis}.
\newblock \penalty0 (1):\penalty0 1--16, 2012.

\bibitem[Kushner(2013)]{kushner2013heavy}
H.~Kushner.
\newblock \emph{Heavy traffic analysis of controlled queueing and communication
  networks}, volume~47.
\newblock Springer Science \& Business Media, 2013.

\bibitem[Lai and Robbins(1985)]{lai1985asymptotically}
T.~L. Lai and H.~Robbins.
\newblock Asymptotically efficient adaptive allocation rules.
\newblock \emph{Advances in applied mathematics}, 6\penalty0 (1):\penalty0
  4--22, 1985.

\bibitem[Lamperti(1960)]{LAMPERTI1}
J.~Lamperti.
\newblock Criteria for the recurrence or transience of stochastic process. i.
\newblock \emph{Journal of Mathematical Analysis and Applications}, 1\penalty0
  (3):\penalty0 314 -- 330, 1960.
\newblock \doi{https://doi.org/10.1016/0022-247X(60)90005-6}.

\bibitem[Lamperti(1963)]{LAMPERTI2}
J.~Lamperti.
\newblock Criteria for stochastic processes ii: Passage-time moments.
\newblock \emph{Journal of Mathematical Analysis and Applications}, 7\penalty0
  (1):\penalty0 127 -- 145, 1963.
\newblock \doi{https://doi.org/10.1016/0022-247X(63)90083-0}.

\bibitem[Lattimore and Szepesv{\'a}ri(2020)]{lattimore2018bandit}
T.~Lattimore and C.~Szepesv{\'a}ri.
\newblock Bandit algorithms.
\newblock \emph{preprint}, 2020.

\bibitem[Mei et~al.(2020)Mei, Xiao, Szepesvari, and Schuurmans]{mei2020global}
J.~Mei, C.~Xiao, C.~Szepesvari, and D.~Schuurmans.
\newblock On the global convergence rates of softmax policy gradient methods.
\newblock \emph{arXiv preprint arXiv:2005.06392}, 2020.

\bibitem[Meyn and Tweedie(2012)]{meyn2012markov}
S.~P. Meyn and R.~L. Tweedie.
\newblock \emph{Markov chains and stochastic stability}.
\newblock Springer Science \& Business Media, 2012.

\bibitem[Seldin et~al.(2012)Seldin, Szepesv{\'a}ri, Auer, and
  Abbasi-Yadkori]{seldin2012evaluation}
Y.~Seldin, C.~Szepesv{\'a}ri, P.~Auer, and Y.~Abbasi-Yadkori.
\newblock Evaluation and analysis of the performance of the exp3 algorithm in
  stochastic environments.
\newblock In \emph{EWRL}, pages 103--116, 2012.

\bibitem[Srikant and Ying(2019)]{srikant2019finite}
R.~Srikant and L.~Ying.
\newblock Finite-time error bounds for linear stochastic approximation and td
  learning.
\newblock \emph{arXiv preprint arXiv:1902.00923}, 2019.

\bibitem[Sutton and Barto(2018)]{sutton2018reinforcement}
R.~S. Sutton and A.~G. Barto.
\newblock \emph{Reinforcement learning: An introduction}.
\newblock MIT press, 2018.

\bibitem[Thompson(1933)]{thompson1933likelihood}
W.~R. Thompson.
\newblock On the likelihood that one unknown probability exceeds another in
  view of the evidence of two samples.
\newblock \emph{Biometrika}, 25\penalty0 (3/4):\penalty0 285--294, 1933.

\bibitem[Williams(1991)]{williams1991probability}
D.~Williams.
\newblock \emph{Probability with Martingales}.
\newblock Cambridge mathematical textbooks. Cambridge University Press, 1991.
\newblock ISBN 9780521406055.

\bibitem[Williams(1992)]{williams1992simple}
R.~J. Williams.
\newblock Simple statistical gradient-following algorithms for connectionist
  reinforcement learning.
\newblock \emph{Machine learning}, 8\penalty0 (3-4):\penalty0 229--256, 1992.

\bibitem[Zhang et~al.(2020)Zhang, Kim, O'Donoghue, and Boyd]{zhang2020sample}
J.~Zhang, J.~Kim, B.~O'Donoghue, and S.~Boyd.
\newblock Sample efficient reinforcement learning with reinforce.
\newblock \emph{arXiv preprint arXiv:2010.11364}, 2020.

\end{thebibliography}
